\documentclass[11pt,twoside,final]{article}

\usepackage{hyperref}
\hypersetup{colorlinks,linkcolor={blue},citecolor={blue},urlcolor={red}} 

\usepackage{fullpage}

\usepackage[mathscr]{eucal}
\usepackage{amsbsy}

\DeclareMathAlphabet{\mathpzc}{OT1}{pzc}{m}{it}

\usepackage{epsf}
\usepackage{fancyheadings}
\usepackage{graphics}
\usepackage{graphicx}
\usepackage{enumitem}
\usepackage{float}

\usepackage{color}

\usepackage{amsthm} \usepackage{amsfonts} \usepackage{amsmath}
\usepackage{amssymb}

\usepackage{algorithm} \usepackage{mathabx} \usepackage{algorithmic}
\usepackage{bbm}

\usepackage{caption} \usepackage{subcaption}

\usepackage[square, numbers]{natbib}

 \usepackage{macros}
\usepackage[noabbrev,nameinlink]{cleveref}

\newcommand{\figdir}{figures/}

\newcommand{\State}{\ensuremath{X}}

\newcommand{\actionspace}{\ensuremath{\mathcal{A}}}

\newcommand{\rsam}{\ensuremath{r}}

\newcommand{\Noise}{\ensuremath{W}}

\begin{document}


\begin{center}

  {\bf{\LARGE{Krylov--Bellman boosting: Super-linear policy evaluation
        in general state spaces}}}
\vspace*{.2in}

{\large{
\begin{tabular}{ccc}
Eric Xia$^{\star, \dagger}$ && Martin J. Wainwright$^{\star, \dagger,
  \ddagger}$
\end{tabular}
}}

\vspace*{.2in}

\begin{tabular}{c}
Department of Statistics$^\star$, and \\ Department of Electrical
Engineering and Computer Sciences$^\ddagger$ \\ UC Berkeley, Berkeley,
CA \\
\end{tabular}

\vspace*{.2in}

\begin{tabular}{c}
  Department of Electrical Engineering and Computer
  Sciences$^\dagger$\\
  Department of Mathematics$^\ddagger$ \\
  Massachusetts Institute of Technology, Cambridge, MA
\end{tabular}

\medskip

\today

\vspace*{.2in}

\begin{abstract}
We present and analyze the Krylov--Bellman Boosting (KBB) algorithm for
policy evaluation in general state spaces.  It alternates between
fitting the Bellman residual using non-parametric regression (as in
boosting), and estimating the value function via the least-squares
temporal difference (LSTD) procedure applied with a feature set that
grows adaptively over time. By exploiting the connection to Krylov
methods, we equip this method with two attractive guarantees.  First,
we provide a general convergence bound that allows for separate
estimation errors in residual fitting and LSTD computation.
Consistent with our numerical experiments, this bound shows that
convergence rates depend on the restricted spectral structure, and are
typically super-linear. Second, by combining this meta-result with
sample-size dependent guarantees for residual fitting and LTSD
computation, we obtain concrete statistical guarantees that depend on
the sample size along with the complexity of the function class used
to fit the residuals.  We illustrate the behavior of the KBB algorithm
for various types of policy evaluation problems, and typically find
large reductions in sample complexity relative to the standard
approach of fitted value iteration.
\end{abstract}

\end{center}


\section{Introduction}

The past decade has witnessed tremendous success with applications of
reinforcement learning (e.g.,~\citep{tobin2017domain, JMLR:v17:15-522,
  silver2016alphago}), but many challenges remain.  Notably, many
procedures in reinforcement learning are very ``data hungry'', meaning
that large sample sizes---which may or may not be available in a given
application---are often required.  Given a data-dependent procedure,
its \emph{sample complexity} is a function that characterizes the
number of samples required, as a function of the problem parameters,
to achieve a desired error tolerance.  Given the cost associated with
sampling (or lack of availability of large sample sizes), an important
problem is to develop RL algorithms with the lowest possible sample
complexity.

In this paper, we focus on the problem of \emph{policy evaluation} for
Markov decision processes defined over general state spaces.  Policy
evaluation occupies a central role in reinforcement learning.  It is
of interest both in its own right and as a key subroutine in many
algorithms.  For instance, classical algorithms for computing optimal
policies, such as policy iteration~\cite{howarddp}, require repeated
rounds of policy evaluation; iterative schemes like policy gradient
methods~\cite{reinforce1992, silver2014policygrad,
  sutton1999policygrad, kakade2001npg} and actor-critic
schemes~\cite{konda2001actorcritic, pmlr-v48-mniha16, wu2017acktr,
  bhatnagar2009naturalac} involve maintaining estimates of the value
function as a part of the algorithm.

Policy evaluation in discrete state spaces, known as the tabular
setting, is very well-understood (e.g.,~\cite{bertsekas2007dp2,
  bellman1957markov, putermanbrumelle1979}); notably, recent work has
exhibited various procedures, for both batch and on-line settings,
that achieve fundamental lower bounds in a strong instance-optimal
sense~\cite{khamaru2020PE, pananjady2021instance}. Far more
challenging are Markov decision processes involving continuous state
spaces.  In this setting, there are at least two broad classes of
approaches.  One approach is based on using a suitably ``rich''
non-parametric function class, such as regression trees, reproducing
kernel Hilbert spaces, or neural networks, to repeatedly refit an
estimate of the value function; procedures of this type are known
collectively as instantiations of fitted value iteration
(FVI)~\cite{JMLR:v6:ernst05a}.  The FVI algorithm along with its close
relative fitted $Q$-iteration (FQI), has been the focus of a long line
of work (e.g.,~\cite{antos2007fittedQ, JMLR:v9:munos08a,
  scherrer2015approximate,chen2019infobatch, pmlr-v120-yang20a,
  duan2020opelin}).

Another classical approach to approximate policy evaluation, even
simpler than fitted value iteration, is via the \emph{least-squares
temporal difference} method, or LSTD for
short~\cite{Bradtke2004LinearLA, Sutton1998,
  boyan2002technical}. Given a finite collection of features or basis
functions, it computes the value function within the associated linear
span that provides the ``best fit'', as formalized in terms of a
projected fixed point (e.g.,~\cite{bertsekas2009projected,
  yu2010projlin}).  There has been line of work analyzing algorithms
for computing the LSTD solution~\cite{tsitsiklis97funcapprox,
  sutton2008offlstd, pmlr-v75-bhandari18a,
  pmlr-v84-lakshminarayanan18a}, as well as more general settings
involving projected fixed-point equations~\cite{MouPanWai20,
  wang2022finitetime, duan2021optimal}. This line of analysis, for the
most part, takes as input a fixed set of features.  In practice, the
quality of the LSTD fit depends strongly on the given basis, and
finding a good basis is often a difficult task.  Often, features are
chosen based on knowledge of the specific problem domain; there is
also a line of past work~\cite{keller2006autobasis, menache2003bfa} on
automated procedures for feature construction.

\subsection{Our contributions}

In this paper, we propose and analyze a procedure for approximate
policy evaluation that combines the LSTD framework with the boosting
approach of fitting residuals.  The method estimates the value
function by computing a sequence of LSTD solutions, where the basis
used at each round is augmented with a non-parametric estimate of the
Bellman residual from the previous round.  The LSTD steps can be
interpreted as computing a sequence of Krylov subspaces, and
accordingly we refer to our method as the
\emph{Krylov--Bellman boosting} algorithm, or KBB for short.

We note that boosting methods---that is, using non-parametric
procedures to fit the Bellman residual---have been studied in past
work on reinforcement learning tasks~\cite{abel2016qboost}.  In our
method, however, the combination with LSTD is equally important.  As
shown in our theoretical analysis, it is the combination of boosting
with LSTD that allows us establish convergence guarantees for KBB that
are provably superior to those of fitted value iteration (FVI).
Consistent with this theory, we also see the gains of KBB over FVI in
our experimental studies. In the tabular setting, an idealized form of
the KBB algorithm with known transition dynamics---and hence no error
in fitting the residuals---has been studied in past
work~\cite{parr2007bebfs, Parr2008AnAO}, in which context it is called
the method of Bellman error basis functions (BEBF).  However for
general state spaces it is unrealistic to assume that residuals can be
computed without error; for continuous state spaces this task can be
challenging even if the model dynamics are known, since it involves
computing the integral that defines the Bellman update.  Even more
challenging---and the focus of this paper---is the setting of unknown
dynamics, in which case any estimate of the residual is subject to
statistical error due to a finite sample size.  It has been
noted~\cite{petrik2007laplace} that the BEBF method is equivalent to a
Krylov subspace method~\cite{saad1981krylov, saad2003linalg} for
solving a finite-dimensional linear system. This connection to Krylov
methods plays an important role in our paper; as opposed to the
classical use of Krylov methods in solving finite-dimensional linear
systems, in this paper we tackle the more general problem of solving a
linear operator equation, which (for general continuous state spaces)
corresponds to an infinite-dimensional problem.

The primary contributions of this paper take the form of two theorems.
Our first main result, stated as~\Cref{thm:main}, is a general
convergence guarantee for the KBB algorithm.  It involves a sequence
of so-called restricted spectral values (cf.~\cref{eqn:ortho-eig})
that track how the effective conditioning of the residual problem
improves as the algorithm proceeds.  This guarantee should be viewed
as a ``meta-result'', since it applies to arbitrary procedures used to
estimate the Bellman residual and to approximate the LSTD solutions,
with the convergence guarantee depending upon these two associated
errors.  In particular, ~\Cref{thm:main} applies both when the
transition dynamics are known and numerical integration is used to
approximate the Bellman residual at each round, and in the
reinforcement learning setting where the dynamics are unknown but
samples from the Markov chain are available.  Our second set of
results, stated as~\Cref{thm:regression} and~\Cref{cor:final},
provides guarantees for this latter RL setting, and in particular, for
independent samples of state, next-state and reward triples.  Under
this set-up, \Cref{thm:regression} provides an upper bound on the
regression error for a general function class, whereas
\Cref{cor:final} combines this guarantee with \Cref{thm:main} to
provide an ``end-to-end'' guarantee for a particular instantiation of
the KBB algorithm.


\subsection{A preview}

\begin{figure}[ht!]
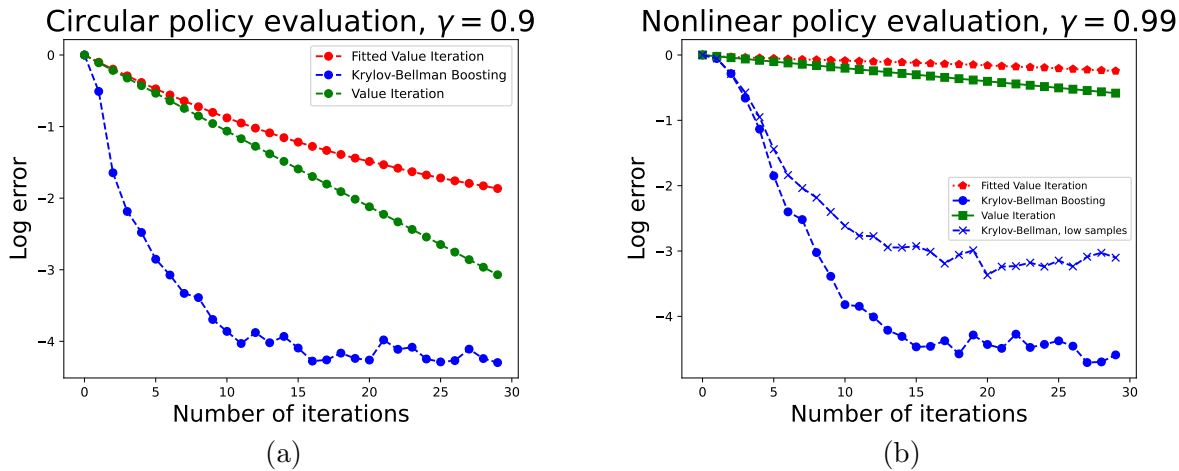

  \begin{center}
    \begin{tabular}{cc}
  \widgraph{0.5\textwidth}{\figdir/circular_gamma0.9} &
  \widgraph{0.5\textwidth}{\figdir/nonlinearwlow_gamma0.99} \\ (a) &
  (b)
    \end{tabular}
    \caption{Plots of log error in $\munorm{\cdot}$-norm versus
      iteration number for three algorithms for policy evaluation:
      exact value iteration with known dynamics (red), fitted value
      iteration (green), and Krylov Bellman Boosting (blue).  (a)
      ``Circular'' corresponds to a MRP arising from a random walk on
      the circle (cf.~\Cref{sec:plots-tabular}) with discount factor
      $\contractPar = 0.90$. (b) A non-linear system
      (cf.~\Cref{sec:plots-NonLinear}) with discount factor
      $\contractPar = 0.99$.}
    \label{fig:initial}
  \end{center}
\end{figure}

In order to whet the reader's appetite, let us provide an illustrative
preview of the results to come.  In~\Cref{fig:initial}, we compare the
performance of three different methods for policy evaluation: the
classical value iteration algorithm (green); a version of fitted value
iteration (FVI) (red), and the KBB algorithm analyzed in this paper
(blue). In our implementations of both KBB and FVI, we used a gradient
boosting procedure to fit the residual at each round (KBB), or the 
value function at each round (FVI).  In each case, we used the same
number of samples for each fit, so that for a fixed iteration number,
both procedures use the same sample size.  In contrast, classical
value iteration is predicated upon exact knowledge of the dynamics,
and computes Bellman updates exactly.

In each panel, we plot the log error in evaluating a given policy
versus the number of iterations.  Panel (a) corresponds to a simple
tabular problem that arises from a random walk on the circle
(see~\Cref{sec:plots-tabular} for details), whereas panel (b)
corresponds to a Markov reward process arising from non-linear
dynamical system with a continuous state space
(see~\Cref{sec:plots-NonLinear} for details).  Consistent with extant
theory, both classical and fitted value iteration converge at a
geometric rate, as shown by the linear nature of these plots of log
error versus iteration number.  In contrast, the KBB algorithm
exhibits super-linear convergence up to an error floor, which arises
from the statistical errors in evaluating residuals, and is consistent
with our theory (cf.~\Cref{cor:final}). In the nonlinear setting, we
have also included the plot of running the KBB algorithm with less
samples, still indicating super-linear convergence with a higher error
floor.

\begin{figure}[h!]
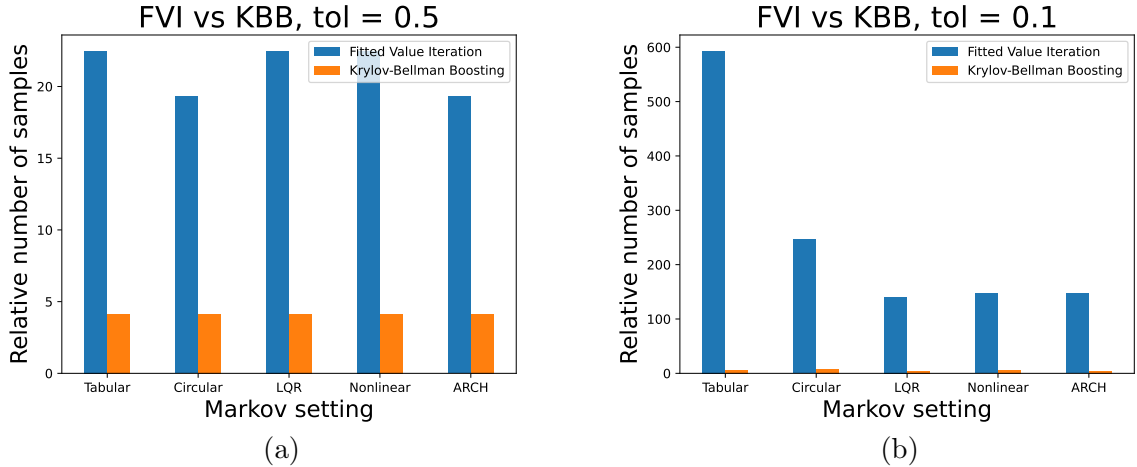

  \begin{center}
    \begin{tabular}{cc}
  \widgraph{0.5\textwidth}{\figdir/0.9_tol0.5_compare} &
  \widgraph{0.5\textwidth}{\figdir/0.9_tol0.1_compare} \\ (a) & (b)
    \end{tabular}
    \caption{ Comparison of the sample complexities of Krylov--Bellman
      boosting (KBB) to fitted value iteration (FVI), as measured by
      the (relative) sample size required to reduce the error by a
      factor of $1/2$ (panel (a)), or $1/10$ (panel (b)).  Both
      algorithms are applied to five different types of Markov reward
      processes, as indicated on the horizontal axis.  For each MRP,
      the bar plots compare the relative number of samples used, with
      the KBB bars rescaled to remain at a constant level across
      problem types.}
    \label{fig:comparison}
  \end{center}
\end{figure}

In practice, of primary importance in choosing between the FVI and KBB
algorithms are their respective sample complexities---that is, how
many samples are required to achieve a specified accuracy?
Figure~\ref{fig:comparison} provides a comparison of the sample
complexities of these two procedures for five different types of
Markov reward processes.  Panels (a) and (b) compare the sample
complexity required to reduce the error by factors of $1/2$ and
$1/10$, respectively.
For each model within each panel, we rescale the KBB sample size so
that it remains constant across models; we apply this same scaling to
the FVI sample size, so the relative heights of the bars are the
relevant comparison.  In panel (a), we see that the KBB procedure
requires roughly a factor of $4$--$5$ times fewer samples than the FVI
procedure.  As shown in panel (b), the savings are even more dramatic
when a higher error tolerance is used.


\subsection{Paper organization and notation}

The remainder of the paper is organized as
follows. Section~\ref{sec:background} is dedicated to the background
necessary to understand our algorithm and results, including Markov
reward process, linear function approximation and the least-squares
temporal difference (LSTD) estimate, along with some background on Krylov
subspace methods for solving linear systems. In
Section~\ref{sec:main-results}, we describe the Krylov--Bellman
boosting (KBB) algorithm, and then state and interpret our two main
guarantees (\Cref{thm:main} and~\Cref{thm:regression}) on its
behavior.  In Section~\ref{sec:plots}, we provide numerical results
that illustrate the behavior of the KBB algorithm in various settings,
along with comparisons to fitted value iteration.
Section~\ref{sec:proofs} is devoted to the proofs of our main results,
with more technical details deferred to the appendices. We conclude
with a discussion in Section~\ref{sec:discussion}.


\paragraph{Notation:} Let us summarize some notation used throughout the
paper.  For a probability distribution $\stationary$, let $\LsqClass$
denote the space of real-valued functions with domain $\stateset$,
square integrable with respect to $\stationary$.  This space is
equipped with the inner product
\begin{align*}
  \statinprod{f}{g} & \defn \int_\stateset f(x) g(x) d \stationary(x)
  \equiv \EE_{\stationary}[f(X) g(X)],
\end{align*}
along with the norm $\munorm{f} = \sqrt{\statinprod{f}{f}}$. Given a
subspace \mbox{$\subspace \subset \LsqClass$,} we write $v
\perp_\stationary \subspace$ to indicate that $v$ is orthogonal to the
$\subspace$ with respect to the inner product
$\statinprod{\cdot}{\cdot}$, and $\subspace^\perp$ to denote the set
of all such vectors $v$. For an operator $\opQ$, we write $\opQ
\succeq 0$ to mean $\statinprod{f}{\opQ f} \geq 0$ \text{for all $f
  \in \LsqClass$,} and we say that $\opQ$ is a \emph{positive
operator} for short.  Given two operators $\TransOp$ and $\opQ$, we
write $\TransOp \succeq \opQ$ to indicate that $\TransOp - \opQ
\succeq 0$. Given samples $\{ \state_i \}_{i=1}^\numobs$, we define
the \emph{empirical norm} $\empnorm{f} \defn \sqrt{\frac{1}{n}
  \sum_{i=1}^\numobs f^2(\state_i)}$; note that it is a random
quantity due to its dependence on the underlying samples.  Throughout
the paper, we use $(c, c', c_1, c_2)$ etc. to denote universal
constants whose values may change from line to line.


\section{Background}
\label{sec:background}

Here we describe some background necessary to explain and motivate our
results. In Section~\ref{sec:mrp} we describe the basic setup of the
problem of interest, solving for the value function of a Markov reward
process, as well as the sampling model to which we have
access. Section~\ref{sec:lstd} then describes an approximate method
for solving for the value function as linear combination of given
basis functions. Finally in Section~\ref{sec:krylov} we describe
Krylov subspace methods for solving linear equations.


\subsection{Markov decision and reward processes}
\label{sec:mrp}

A Markov decision process consists of a state space $\stateset$, an
action space $\actionspace$, a reward function $\reward$, along with a
collection of probability transition functions.  In this paper, we
focus exclusively on the policy evaluation procedure, in which case a
given policy is fixed up front.  For a fixed policy, a Markov decision
process reduces to a \emph{Markov reward process} (MRP), which can be
characterized more simply in terms of the state space $\stateset$,
along with a probability transition kernel $\Trans$, along with a
reward function.

In a Markov reward process, the states evolve over time according to a
collection of transition kernels $\Trans$.  At some time $t = 1, 2,
\ldots$, when in state $\state_t$, the next state is generated by
sampling $\State_{t+1} \sim \Trans(\cdot \mid \state_t)$, where for
each $\state \in \stateset$, the function $\Trans(\cdot \mid \state)$
defines a probability measure over $\stateset$.  Under standard
regularity assumptions, the operator $\Trans$ has a stationary measure
$\stationary$ satisfying the consistency condition
\begin{align*}
\stationary(\state') = \EE_{\State \sim \stationary}[\Trans(\state'
  \mid \State)].
\end{align*}
The MRP also is equipped with a reward function $\reward: \stateset
\to \RR$, which we assume to be uniformly bounded and
$\stationary$-measurable.  At time $t = 1, 2, \ldots$, when in state
$\state_t$ we receive a reward $\reward(\state_t)$.

The problem of policy evaluation corresponds to computing the
\emph{value function}
\begin{align}
\valuestar(\state) \defn \EE\big[ \sum_{t=0}^\infty \contractPar^t
  \cdot \reward(\State_t) \mid \state_0 = \state \big],
\end{align}
where the conditional expectation is taken over a trajectory
$(\State_1, \State_2, \ldots)$ from the underlying Markov chain. Since
the reward function $\reward$ is measurable and uniformly bounded, the
value function exists and is uniquely defined.  Moreover, it is
measurable, uniformly bounded by $\frac{\inftynorm{\reward}}{1 -
  \contractPar}$, and therefore $\valuestar \in \LsqClass$.

Given a value function $\values \in \LsqClass$ we define the Bellman
operator $\bellmanOp: \LsqClass \to \LsqClass$ via
\begin{align}
\label{eqn:bellman-fixed}
\bellmanOp(\values)(\state) \defn \reward(\state) + \contractPar
\EE_{\state' \sim \Trans(\cdot \mid \state)}[\values(\state')] \qquad
\text{for all $\state \in \stateset$}.
\end{align}
By classical results on dynamic programming, the value function is a
fixed point of the Bellman operator---that is, $\valuestar =
\bellmanOp(\valuestar)$.  Moreover, the Bellman operator $\bellmanOp$
is $\discount$-contractive in the norm $\munorm{\cdot}$; see the
standard references~\cite{puterman2014markov, Sutton1998,
  Bertsekas2009} for further background.

In order to clarify the connection to solving linear operator
equations, define the transition operator $\TransOp(\values)(\state)
\defn \EE_{\state' \sim \Trans(\cdot \mid \state)} \big[
  \values(\state') \big]$.  Observing that $\TransOp$ is a linear
operator on $\LsqClass$, we can rewrite the Bellman fixed point
relation as
\begin{align}
\label{eqn:linearop-formulation}
\valuestar = \reward + \contractPar \TransOp \valuestar \iff (\IdOp -
\contractPar \TransOp) \valuestar = \reward,
\end{align}
where $\IdOp$ denotes the identity operator on $\LsqClass$.  Thus,
computing the value function is equivalent to solving the linear
operator equation~\eqref{eqn:linearop-formulation}.

We conclude by describing the sampling model studied in this paper.
The operator $\TransOp$ and the reward function $\reward$ are unknown
to us, but we have access to sample transition pairs $(\state, \rsam,
\state')$ generated as follows
\begin{align}
\label{eqn:sample-model}
\state \sim \stationary, \quad \rsam = \reward(\state), \quad
\text{and} \quad \state' \sim \Trans(\cdot \mid \state).
\end{align}
To be concrete, intermediate stages of our procedure make use of a
dataset $\mathcal{D} = \{(\state_t, \rsam_t,
\state'_t)\}_{t=1}^\numobs$ of $\numobs$ samples.


\subsection{Linear function approximation and LSTD}
\label{sec:lstd}

When the state space $\stateset$ is sufficiently complex---either
finite but with a very large number of states, or continuous in
nature---any efficient scheme for policy evaluation requires some form
of function approximation. In loose terms, we choose some function
space, and then seek to find the ``best'' approximation to the value
function within this space.

A classical and widely-used approach is based on \emph{linear function
approximation.}  Given a collection of functions $\basis_j \in
\LsqClass$ for $j = 1, \ldots, J$, known either as features or basis
functions, we use the subspace $\subspace = \myspan\{ \basis_1,
\basis_2, \ldots, \basis_J \}$ to approximate the value function.  An
attractive feature is that any function in the subspace $\subspace$
can be written as $V_\coeffs = \sum_{j=1}^J \coeffs_j \basis_j$ for
some vector $\coeffs \in \RR^J$, so that computations can be reduced
to linear-algebraic operations over $\RR^J$.  One way to define the
best approximation is via the notion of a projected fixed point
(e.g.,~\cite{bertsekas2009projected, yu2010projlin}).  Defining the
projection onto the subspace $\subspace$ via
\begin{align*}
\Proj(f) \defn \arg \min_{\values \in \subspace} \munorm{f - \values},
\end{align*}
we seek a function $\valuesLSTD \in \subspace$ that satisfies the
projected fixed point condition
\begin{align}
\label{eqn:proj-fixed}
\valuesLSTD = \Proj\big( \bellmanOp (\valuesLSTD) \big).
\end{align}
Since $\Proj$ is non-expansive and $\bellmanOp$ is a contraction, both
with respect to the $\munorm{\cdot}$-norm, there is a unique solution
$\valuesLSTD$, known as the \emph{population LSTD solution}.  For
future reference, we note that an abstract characterization of
$\valuesLSTD$ is in terms of the orthogonality conditions
\begin{align}
\label{eqn:lstd-ortho}
\statinprod{\basis_j}{\bellmanRes(\valuesLSTD)} & = 0 \qquad \text{for
  all $j = 1, \ldots, J$,}
\end{align}
where $\bellmanRes(\valuesLSTD) \defn \valuesLSTD -
\bellmanOp(\valuesLSTD)$ is the \emph{Bellman residual}.

Less abstractly, the computation of the LSTD estimate $\valuesLSTD$
can be reduced to solving a linear system over $\real^J$.  Let
\mbox{$\basisVec(\state) = (\basis_1(\state), \basis_2(\state),
  \ldots, \basis_t(\state)) \in \RR^t$} be the vector obtained by
evaluating each basis function at the state $\state$. By standard LSTD
theory~\cite{Bradtke2004LinearLA}, we can compute the LSTD solution
$\valuesLSTD$ by solving for $\coeffsLSTD \in \real^t$ in the linear
system
\begin{align*}
\EE_{(\State, \State')}\big[ \basisVec(\State) \big( \basisVec(\State)
  - \contractPar \basisVec(\State') \big)^T \big] \coeffsLSTD =
\EE_{\State \sim \stationary} \big[ \reward(\State) \basisVec(\State)
  \big],
\end{align*}
and then setting $\valuesLSTD = \values_{\coeffsLSTD}$.

In most practical settings, we cannot compute these expectations
exactly, but instead have access to samples $\data = \{(\state_i,
\rsam_i, \state'_i)\}_{i=1}^\numobs$ generated according to the
model~\eqref{eqn:sample-model}.  We can use these samples to form
empirical estimates as follows. The \emph{plug-in estimate}
$\coeffsLSTDest$ is obtained by solving the linear system
\begin{align}
  \label{EqnEmpiricalLSTD}
\big( \frac{1}{\numobs} \sum_{i=1}^\numobs \basisVec(\state_i)
(\basisVec(\state_i) - \contractPar \basisVec(x_i'))^T \big)
\coeffsLSTDest = \frac{1}{\numobs} \sum_{i=1}^\numobs \rsam_i
\basisVec(\state_i),
\end{align}
which leads to the \emph{empirical LSTD estimate} $\valuesLSTDest =
\values_{\coeffsLSTDest}$.


\subsection{Krylov subspace methods}
\label{sec:krylov}

Portions of our development rely on connections to Krylov subspace
methods for solving systems of linear equations, which we describe
briefly here.  See the book by Saad~\cite{saad2003linalg} for more
details.  Suppose we are interested in solving a linear system $\matA
x = \vecb$, where $\matA \in \RR^{d \times d}$ is a matrix, and $\vecb
\in \RR^n$ is a vector. In many cases, the dimension $d$ may be very
large so directly solving the system of equations is computationally
very expensive. Krylov subspace methods are based on approximating the
solution to such a linear system via an expanding sequence of
subspaces.  For each positive integer $j = 1, \ldots, d$, we define
the $j^{th}$-order Krylov subspace as
\begin{align}
\label{EqnKrylovSubspace}
\krylov_j(\matA, \vecb) \defn \text{span}\big\{\vecb, \matA \vecb,
\matA^2 \vecb, \ldots, \matA^j \vecb \big\}.
\end{align}
Using the subspace $\krylov_j(\matA, \vecb)$, we can define an
approximate solution $\widehat{x} \in \krylov_j(\matA, \vecb)$
\begin{align}
\label{eqn:orth-proj-process}
\widehat{x} \in \krylov_j(\matA, \vecb) \qquad \text{such that} \qquad
\vecb - \matA \widehat{x} \perp \krylov_j(\matA, \vecb).
\end{align}
Many algorithms for computing approximations to $x$ involve forming
the Krylov subspace $\krylov_j(\matA, \vecb)$ and then solving the
system~\eqref{eqn:orth-proj-process}; the conjugate gradient method is
a notable instance.  In the setting of tabular reinforcement learning
the problem of policy evaluation is equivalent to solving a linear
system; thus Krylov methods are applicable, as has been noted in
past work~\cite{petrik2007laplace}.  Our method, to be described in
the next section, exploits the connection to Krylov theory at the more
abstract level of operators, and allows for statistical errors in the
updates.


\section{Krylov--Bellman boosting and its guarantees}
\label{sec:main-results}

In this section, we first describe the Krylov--Bellman boosting (KBB)
procedure (\Cref{sec:alg-descrip}), and then state some theoretical
guarantees on its performance (\Cref{sec:conv-statements}).  The KBB
algorithm involves a sequence of LSTD solutions, along with a sequence
of fits to the Bellman residual. Our first result (\Cref{thm:main}) is
a general guarantee that allows for arbitrary errors in these
intermediate computations.  In~\Cref{sec:regression}, we analyze a
general family of non-parametric procedures for fitting the residual,
and provide an end-to-end result that bounds the behavior of the KBB
procedure with statistical error.


\subsection{Krylov--Bellman Boosting}
\label{sec:alg-descrip}

In operational terms, the KBB procedure can be viewed as a procedure
for adaptively choosing features or basis function; each feature is a
function belonging to $\LsqClass$.  At each iteration $t = 1, 2,
\ldots$, the KBB procedure maintains a collection of features
$\{\basis_j\}_{j=1}^t$ that defines the subspace $\subspace_t =
\text{span} \big \{ \basis_1, \ldots, \basis_t \}$.  In the idealized
(but unimplementable) form of the procedure, there are two steps
involved in the update from iterate $t$ to $t+1$:
\begin{itemize}
\item Compute the population level LSTD
  solution~\eqref{eqn:proj-fixed} defined by the subspace
  $\subspace_t$.  Call it $\valuesLSTD_t$.
\item Compute the Bellman residual $\bellmanRes(\valuesLSTD_t)$ and
  use it to form the augmented subspace
\begin{align*}
\subspace_{t + 1} = \text{span}\big\{ \basis_1, \basis_2, \ldots,
\basis_t, \bellmanRes(\valuesLSTD_t) \big \}.
\end{align*}
\end{itemize}

Suppose that we initialize this procedure with the subspace
$\subspace_1 = \text{span} \{\reward \}$, corresponding to the span of
the reward function.  With this choice, the subspace $\subspace_{t+1}$
is equivalent to the Krylov subspace~\eqref{EqnKrylovSubspace} of
order $t + 1$ defined by the linear operator $\opQ = \IdOp -
\contractPar \TransOp$ and the reward function $\reward$---more
specifically, we have
\begin{align*}
\subspace_{t+1} = \krylov_{t+1}(\opQ, \reward) \equiv \text{span} \big
\{ \reward, \opQ(\reward), \opQ^2(\reward), \ldots, \opQ^{t}(\reward)
\big \}.
\end{align*}
Moreover, observe that the orthogonality
condition~\eqref{eqn:lstd-ortho} that defines the population LSTD
solution is equivalent to the orthogonality
condition~\eqref{eqn:orth-proj-process} that defines the associated
Krylov subspace.  Thus, this idealized algorithm is exactly an
abstract Krylov subspace method for computing the solution
$\valuestar$ to the linear operator equation $(\IdOp - \contractPar
\TransOp) \valuestar = \reward$ over the space $\LsqClass$.

In the setting of unknown model dynamics and general state spaces,
 this idealized scheme is unimplementable for two
reasons.  First, the population LSTD solution cannot be computed
exactly; instead we must approximate it, typically with the empirical
LSTD solution~\eqref{EqnEmpiricalLSTD}.  A second, more daunting,
challenge is the computation of the Bellman residual at each round.

In the Krylov--Bellman boosting procedure, we assume access to a
mechanism that provides data of the form $(\state_i, \rsam_i,
\state'_i)$ generated according to the sampling
model~\eqref{eqn:sample-model}, with which we estimate the desired
quantities. As described in Section~\ref{sec:lstd}, a dataset of the
form $\LSTDdata = \{(\state_j, \rsam_j, \state'_j)\}_{j=1}^m$ can be
used to compute the empirical LSTD
estimate~\eqref{EqnEmpiricalLSTD}. Moreover, given our current
estimate $\values$ of the value function, consider the problem of
estimating the Bellman residual~$\bellmanRes(\values)$ using a dataset
$\REGdata = \{(\state_i, \rsam_i, \state'_i)\}_{i=1}^n$.  In the
specific instantiation of the KBB algorithm given here, we implement
and analyze a \emph{non-parametric least-squares estimate} of this
residual.  More precisely, given a suitably chosen function class
$\fclass$ and the dataset $\REGdata$, we compute the approximation
\begin{align}
\label{eqn:regression}
\bellmanRes(\values) \approx \arg \min_{f \in \fclass} \Big \{
\frac{1}{n} \sum_{i=1}^n \Big [ \values(\state_i) - \big( \rsam_i +
  \contractPar \values(\state_i') \big) - f(\state_i) \Big ]^2 \Big
\}.
\end{align}
The Krylov--Bellman boosting (KBB) procedure alternates between LSTD
fitting and fitting of the residual, as stated formally as
Algorithm~\ref{KBB}.
\begin{varalgorithm}{KBB}
\caption{Krylov--Bellman Boosting}
\label{KBB}
\begin{algorithmic}[1]
\STATE \texttt{Inputs:} (i) Datasets $\LSTDdata_t$ and $\REGdata_t$, $
t= 0, 1, 2, \ldots$, each containing transition tuples generated
according to the model~\eqref{eqn:sample-model}. (ii) Function class
$\fclass$ over which to perform regression.
\vspace{6pt}

\STATE Initialize $\values_0 = 0$, $\basisSet_0 = \emptyset$.

\FOR {$t = 0, \ldots$}

\STATE Use dataset $\REGdata_{t} = \{ (\state_i, \rsam_i, \state_i')
\}_{i=1}^{\Rnumobs_t}$ to estimate $\bellmanRes(\values_{t})$ by
solving the non-parametric least-squares problem:
\begin{align}
\label{EqnAlgRegression}  
\basis_{t+1} & \in \arg \min_{f \in \fclass} \left \{
\frac{1}{\Rnumobs_{t}} \sum_{i=1}^{\Rnumobs_{t}} \big [
  \values_{t}(\state_i) - \big(\rsam_i + \contractPar \cdot
  \values_{t}(\state_i') \big) - f(\state_i) \big ]^2 \right \}.
\end{align}

\STATE Update basis set $\basisSet_{t+1} = \basisSet_{t} \cup \{
\basis_{t+1} \}$. \\
Define function \mbox{$\state \mapsto \basisVec_{t+1}(\state) \defn
  \big( \basis_1(\state), \basis_2(\state), \ldots,
  \basis_{t+1}(\state) \big) \in \RR^{t+1}$.}

\STATE Use dataset $\LSTDdata_{t} = \{ (\state_i, \rsam_i, \state_i')
\}_{i=1}^{\Lnumobs_t}$ to estimate LSTD coefficient vector
$\coeffs^{t+1} \in \RR^{t+1}$ via
\begin{align*}
\coeffs^{t+1} = \Big[ \frac{1}{\Lnumobs_t} \sum_{i=1}^{\Lnumobs_t}
  \basisVec_{t+1}(\state_i)(\basisVec_{t+1}(\state_i) - \contractPar
  \basisVec_{t+1}(\state_i'))^T \Big]^{-1} \; \Big \{
\frac{1}{\Lnumobs_t} \sum_{i=1}^{\Lnumobs_t} \rsam_i \cdot
\basisVec_{t+1}(\state_i) \Big \}.
\end{align*}

\STATE Set $\values_{t+1} = \sum_{j=1}^{t+1} \coeffs^{t+1}_j
\basis_j$.

\ENDFOR

\end{algorithmic}
\end{varalgorithm}

Let us discuss a few choices in how Algorithm~\ref{KBB} can be
implemented.  The first choice concerns how the datasets $\LSTDdata_t$
and $\REGdata_t$ are generated.  In the simplest possible setting,
they are mutually independent for all $t = 0, 1, 2, \ldots$, and each
data set contains transition tuples $(\state, \rsam, \state')$ that
are also generated i.i.d.  More generally, we can form datasets of
this type by collecting triples from trajectories of the Markov
process, in which case the associated dependence needs to be addressed
in any statistical analysis.

The second important choice is the function class $\fclass$ used to
compute the regression estimates~\eqref{EqnAlgRegression}. There are
various families that are commonly used in machine learning and
statistics, including splines and other linear smoothing
methods~\cite{wahba1990spline, hastie01statisticallearning},
reproducing kernel Hilbert spaces~\cite{hofmann2008kernel}, random
forests and regression trees~\cite{breiman1984classification,
  breiman2001random}, boosting procedures~\cite{FREUND1997119}, and
neural networks~\cite{SCHMIDHUBER2015}. For our numerical experiments
presented in Section~\ref{sec:plots}, we used a regression tree
boosting procedure~\cite{chen2016xgboost}.


\subsection{Convergence guarantees}
\label{sec:conv-statements}

We now provide some theoretical guarantees for the Krylov--Bellman
Boosting procedure.  Our first result (\Cref{thm:main}) is of a
general nature: it allows arbitrary procedures to be used in computing
the LSTD solutions and residual fits at intermediate stages of the
algorithm.  In fact, while we have described the KBB residual fits in
terms of non-parametric regression, \Cref{thm:main} actually applies
to other procedures that might be used for this task.  (For instance,
if the model dynamics were known, one could use numerical procedures
to approximate the Bellman update.)  Our second main result
(\Cref{thm:regression}) is more specific in nature, in that it
provides bounds on the error in the LSTD and residual fitting steps
induced by using empirical samples, as used in our introduction of the
procedure.

For ease of notation, define the discount operator $\opQ \defn \IdOp -
\contractPar \TransOp$.  In the analysis given here, we assume that
the Markov chain is reversible.\footnote{ Although our theory exploits
reversibility, it does not seem essential to the practical behavior of
the algorithm itself, as shown by the simulations to follow in the
sequel.}  This assumption implies that the transition operator is a
self-adjoint operator on $\LsqClass$, so that
\mbox{$\statinprod{\TransOp f}{g} = \statinprod{f}{\TransOp g}$} for
all $f, g \in \LsqClass$. This self-adjoint operator $\opQ$ defines
the inner product
\begin{align}
\qinprod{f}{g} \defn \statinprod{f}{\opQ g}, \quad \mbox{along with
  the induced norm } \Qnorm{f} \defn \sqrt{\statinprod{f}{\opQ f}}.
\end{align}
Under certain regularity conditions, it can be shown
(cf.~\Cref{sec:proof-lemma-operator}) that the operator $\opQ \defn
\IdOp - \contractPar \TransOp$ is self-adjoint, invertible, and
satisfies the relation
\begin{align}
\label{eqn:operator}
  (1 - \contractPar) \munorm{f}^2 \leq \statinprod{f}{\opQ f} \leq (1
+ \contractPar) \munorm{f}^2 \qquad \text{for all $f \in \LsqClass$.}
\end{align}
In addition, the operator $\opQ$ is bounded and (under mild regularity
conditions) has a discrete spectrum.

Let
$\REGerr_{t} \defn \basis_{t} - \bellmanRes(\values_{t - 1})$ denote
the error in the regression procedure at round $t$.  Our first
assumption provides control on the regression accuracy:
\myassumption{$\quad$Reg-Err:}{ass:regacc}{For each iteration $t = 1,
  2, \ldots$, the regression procedure is $\REGMSE_t$-accurate:
  \begin{align}
\label{EqnREGvar}    
\tag{RE$(\REGMSE_t)$} \EE \big[ \Qnorm{\REGerr_t}^2 \big] \leq
\REGMSE_t^2.
\end{align}
}

\noindent Our second assumption provides control on the accuracy of
the LSTD computation at each step:
\myassumption{$\quad$LSTD-Err:}{ass:LSTD-err}{At each iteration $t$,
  the approximate LSTD solution $\values_t$ is $\LSTDerr$-accurate:
  \begin{align}
\label{EqnLSTDerr}
\tag{LE$(\LSTDerr_t)$} \Qnorm{\values_t - \valuestar}^2
\stackrel{(a)}{\leq} \Qnorm{\valuesLSTD_t - \valuestar}^2 +
(\LSTDerr)^2, \quad \text{and} \quad \Qnorm{\bellmanRes(\values_t) -
  \bellmanRes(\valuesLSTD_t)} \stackrel{(b)}{\leq} \LSTDerr.
\end{align}
}

Our theory involves the behavior of the operator $\opQ$ when
restricted to certain subspaces of the function space.  Recall that at
each iteration, the algorithm performs LSTD over the linear subspace
$\subspace_t = \text{span}\{ \basis_1, \ldots, \basis_t \}$.  Letting
$\subspace_t^\perp$ denote the orthogonal complement, we define the
constraint set \mbox{$\Constraint_t \defn \big \{ z
  \in \subspace_t^\perp \mid \statinprod{z}{\opQ^{-1} z} = 1 \big
  \}$}, and the \emph{$\Constraint_t$-restricted spectral values}
\begin{align}
\label{eqn:ortho-eig}
\mineig_t \defn \inf_{z \in \Constraint_t^\perp} \, \statnorm{z}^2
\quad \mbox{and} \quad \maxeig_t \defn \sup_{z \in
  \Constraint_t^\perp} \, \statnorm{z}^2.
\end{align}
Note we have the sandwich relation $1 - \contractPar \leq \mineig_t
\leq \maxeig_t \leq 1 + \contractPar$ at each iteration $t$. We let 
$\EE_{t+1}$ denote the expectation over the noise in round $t + 1$.
\begin{theorem}[General KBB bound]
\label{thm:main}
Suppose that the Krylov--Bellman procedure is run with a
$\REGMSE_t$-accurate regression procedure (cf.\ref{EqnREGvar}), and an
$\LSTDerr$-accurate LSTD implementation (cf.~\ref{EqnLSTDerr}).  Then
at each step $t = 1, 2, \ldots$, the error satisfies the $\opQ$-norm
bound
\begin{align*}
\EE_{t+1} \Qnorm{\values_{t+1} - \valuestar}^2 \leq \left(1 -
\tfrac{\mineig_t^2}{8 \maxeig_t} \right) \Qnorm{\values_{t} -
  \valuestar}^2 + \tfrac{10}{\maxeig_t} \cdot \REGMSE_t^2 +
\tfrac{8\maxeig_t}{\mineig_t^2} \cdot (\LSTDerr_t)^2.
\end{align*}
\end{theorem}
\noindent See Section~\ref{SecThmMainProof} for the proof.

\medskip

Note that the function $t \mapsto \frac{\mineig_t^2}{8\maxeig_t}$ is
non-decreasing in $t$ by definition, so convergence is at least
geometric, and can be faster if the ratio
$\frac{\mineig_t^2}{8\maxeig_t}$ is growing quickly. Using the fact
that $\mineig_t \geq 1 - \contractPar$ and $\maxeig_t \leq 1 +
\contractPar$, we have the contraction factor is at least $1 -
\frac{\mineig_t^2}{8\maxeig_t} \leq 1 - \frac{(1 -
  \contractPar)^2}{8(1 + \contractPar)}$ at every iteration. However
in the next section we illustrate that the convergence can be much
faster than this worst case behavior. In addition, we can control the
error terms $\REGMSE_t^2$ and $(\LSTDerr_t)^2$ by the number of
samples, $\Rnumobs_t$ and $\Lnumobs_t$, used in the regression
procedure and LSTD estimate, respectively.

\paragraph{Some intuition:}  Roughly speaking,
the idea underlying the proof is that one round of the KBB algorithm
is at least as good as a gradient-type update that is \emph{restricted
to the subspace} $\Constraint_t^\perp$.  For this reason, the
algorithm depends on the restricted spectral
values~\eqref{eqn:ortho-eig} over this subspace, which determine the
conditioning of the effective problem at each round.

In more detail, suppose both the LSTD and regression procedures are
exact, and consider the problem of minimizing the functional
$\mathcal{F}(\values) \defn \statinprod{\values - \valuestar}{\opQ
  (\values - \valuestar)}$.  It has (Frechet) derivative $\nabla
\mathcal{F}(\values) = \opQ \big(\values - \valuestar \big)$, and
moreover, since the operator $\opQ$ is positive and self-adjoint, this
is a convex optimization problem which achieves its (unique) minimum at
the true value function $\valuestar$.  Using these properties along
with conditions defining LSTD optimality, we can prove
(cf.~\Cref{lems:lstd-minimizer}) that, for any choice of stepsize
$\update > 0$, we have
\begin{align*}
\Qnorm{\values_{t+1} - \valuestar}^2 \leq \Qnorm{\values_t - \update
  \opQ(\values_t - \valuestar) - \valuestar}^2 \qquad
\end{align*}
Using the fact that $\opQ(\values_t - \valuestar) =
\bellmanRes(\values_t)$ and then expanding the square yields
\begin{align*}
\Qnorm{\values_t - \update \bellmanRes(\values_t) - \valuestar}^2 &=
\Qnorm{\values_t - \valuestar}^2 - 2 \update \qinprod{\values_t -
  \valuestar}{\bellmanRes(\values_t)} + \update^2
\Qnorm{\bellmanRes(\values_t)} \\ &\stackrel{(i)}{=} \Qnorm{\values_t
  - \valuestar}^2 - 2\update \statnorm{\bellmanRes(\values_t)}^2 +
\update^2 \Qnorm{\bellmanRes(\values_t)}^2,
\end{align*}
where step (i) makes use of the equivalence $ \opQ(\values_t -
\valuestar) = \bellmanRes(\values_t)$. Since $\values_t$ is the LSTD
fit over $\subspace_t$, the optimality conditions for the projected
fixed point equation~\eqref{eqn:proj-fixed} ensure that
$\bellmanRes(\values_t) \perp \subspace_t$, or equivalently
$\bellmanRes(\values_t) \in \subspace_t^\perp$. As an important
consequence, we can control $\statnorm{\bellmanRes(\values_t)}$ and
$\Qnorm{\bellmanRes(\values_t)}$ in terms of the restricted spectral
values~\eqref{eqn:ortho-eig}. The full proof in~\Cref{SecThmMainProof}
follows this outline.


\subsection{Regression bounds and consequences for KBB}
\label{sec:regression}

Here we provide some results on the regression procedure in
Algorithm~\ref{KBB} for completeness. Given data points $\REGdata = \{
(\state_i, \rsam_i, \state_i')\}_{i=1}^\numobs$, we can write the
regression problem in the generative form as $y_i = \fstar(\state_i) +
\noise_i$ \text{for $i = 1, 2, \ldots, \numobs$}, where
\begin{align}
\label{eqn:generative-model}
\fstar(\state) = \values(\state) - \bellmanOp \values(\state), \qquad
\text{and} \qquad \noise_i = \contractPar \left(\TransOp
\values(\state_i) - \values(\state_i') \right).
\end{align}
Here we have used the fact that the rewards are deterministic. It
should be noted that the ``noise'' variables $\noise_i$ are actually
dependent on the states $\state_i$, so some care is required in
the analysis.

We can write our regression procedure as
\begin{align*}
\fhat \in \arg\min_{f \in \fclass} \frac{1}{n} \sum_{i=1}^\numobs
\left(\response_i - f(\state_i) \right)^2.
\end{align*}
Note that it need \emph{not} be the case that $\fstar \in \fclass$.
In order to allow for the possibility of such mis-specification, we
define the $\statnorm{\cdot}$-projection $\Proj_\fclass$ from
$\LsqClass$ onto $\fclass$, which is well-defined when $\fclass$ is
closed and convex.

Our results involve a non-asymptotic upper bound on $\statnorm{\fhat -
  \bellmanRes(\values)}$; it is easy to convert it to a
$\Qnorm{\cdot}$-guarantee via~\Cref{lems:operator} at the cost of a
$(1 + \contractPar)$- factor. Additionally, it involves a
\emph{statistical estimation error} $\critradius$ that depends on the
function class $\fclass$.  This quantity decreases to zero as the
sample size $\numobs$ increases, but the rate of decrease depends on
the complexity of the function class $\fclass$;
see~\Cref{sec:reg-further-details} for its precise definition, and
associated details.

\begin{theorem}[Regression bound]
  \label{thm:regression}
  For a given function $\values \in \LsqClass$, suppose that we
  compute an estimate $\fhat$ of the Bellman residual
  $\bellmanRes(\values)$ via the regression
  procedure~\eqref{eqn:regression} over a $\bound$-bounded and convex
  function class $\fclass$.  Then we have
  \begin{align}
\EE \munorm{\fhat - \bellmanRes(\values)}^2 \leq c \Big \{
\critradius^2 + \frac{1}{\numobs} \left( \bound^2 +
\inftynorm{\values}^2 \right) +
\munorm{\Proj_\fclass(\bellmanRes(\values)) - \bellmanRes(\values)}^2
\Big \},
\end{align}
  where the statistical estimation error $\critradius^2$ depends on
  the complexity of $\fclass$.
\end{theorem}
\noindent
See Section~\ref{sec:regthm-proof} for the proof.\\

Note that our regression procedures depends on $\critradius^2$, as
well as the approximation error
$\munorm{\Proj_\fclass(\bellmanRes(\values)) -
  \bellmanRes(\values)}^2$; such dependence is to be expected since we
are fitting $\fhat$ in some function class $\fclass$ that does not
necessarily contain $\bellmanRes(\values)$. \\

\noindent Combining~\Cref{thm:main,thm:regression} yields the
following ``end-to-end'' guarantee on the KBB algorithm:
\begin{corollary}[End-to-end guarantee for KBB]
\label{cor:final}  
Under the assumptions of~\Cref{thm:main,thm:regression}, there are
universal constants $c_1$ and $c_2$ such that the KBB update satisfies
\begin{multline*}
\EE_{t+1} \Qnorm{\values_{t+1} - \valuestar}^2 \leq \left(1 -
\tfrac{\mineig_t^2}{8\maxeig_t} \right) \Qnorm{\values_t -
  \valuestar}^2 + \tfrac{c_2 \maxeig_t}{\mineig_t^2} \cdot
(\LSTDerr)^2 \\
+ \frac{c_1}{\maxeig_t} \Big \{ \critradius^2 + \tfrac{1}{\numobs}
\left( \bound^2 + \inftynorm{\values_t}^2\right) +
\munorm{\Proj_\fclass(\bellmanRes \values_t) - \bellmanRes
  \values_t}^2 \Big \},
\end{multline*}
where $\numobs = \Rnumobs_t = |\REGdata_t|$.
\end{corollary}


\subsection{Estimation error via localized complexities}
\label{sec:reg-further-details}

In this section, we provide a detailed description of the statistical
estimation error $\critradius^2$ that appears
in~\Cref{thm:regression}.

 
\subsubsection{Complexity measures and critical radii}

The statistical estimation error can be decomposed as the sum
\begin{align}
\label{EqnDefnCrit}  
\critradius^2 = \empcrit^2 + \popgausscrit^2,
\end{align}
where each of the two quantities $(\empcrit, \popgausscrit)$ arise
from localized complexity measures, in particular sub-Gaussian or
Rademacher complexities, that play a central role in empirical process
theory~\cite{vandeGeer,bartlett05local,dinosaur2019}.  When fitting
the residual $\fstar = \bellmanRes(\values)$, the relevant set is the
\emph{shifted function class} $\shiftfclass = \{f - \ftil \mid f \in
\fclass\}$, where $\ftil \defn \Proj_\fclass(\fstar)$ denotes the
$\statnorm{\cdot}$-projection of $\fstar$ onto $\fclass$.

In the non-parametric fit of the Bellman residual, there are two
sources of ``noise''.  The first is the Bellman noise, and it leads to
the \emph{sub-Gaussian complexity} at scale $\delta > 0$ defined as
\begin{align}
\label{eqn:subgausspop-def}
\subgausspop_\numobs(\delta; \shiftfclass) \defn \EE \left[ \sup_{g \,
    \in \shiftfclass, \munorm{g} \leq \delta} \left| \frac{1}{\numobs}
  \sum_{i=1}^\numobs \noise_i g(\state_i) \right| \right].
\end{align}
Here the $\state_i$ is generated i.i.d.~from the stationary
distribution $\stationary$, whereas the noise $\noise_i$ is generated
according to the model~\eqref{eqn:generative-model}.

The second form of ``noise'' has to do with the discrepancy betwen the
empirical norm $\empnorm{f} \defn \sqrt{\frac{1}{n} \sum_{i=1}^\numobs
  f^2(\state_i)}$ and the population norm $\statnorm{f} =
\sqrt{\Exs[f^2(X)]}$.  In order to prove a uniform bound on this
discrepancy, we make use of the \textit{local Rademacher complexity}
of $\shiftfclass$ at scale $\delta > 0$ as
 \begin{align}
\radcomp_\numobs(\delta; \shiftfclass) \defn \EE \left[ \sup_{g \, \in
    \shiftfclass, \munorm{g} \leq \delta} \left| \tfrac{1}{\numobs}
  \sum_{i=1}^\numobs \rad_i g(\state_i) \right| \right].
 \end{align}
Here $\{\state_i\}_{i=1}^\numobs$ are i.i.d.~samples from the
stationary distribution $\stationary$, and $\{\rad_i\}_{i=1}^\numobs$
are i.i.d.~Rademacher random variables taking values in $\{-1, +1\}$
with equal probabilities, independent of $\{\state_i\}_{i=1}^{\numobs}$.

The two quantities $\popgausscrit$ and $\empcrit$ in the
decomposition~\eqref{EqnDefnCrit} are both quantities known as
\emph{critical radii} in empirical process theory.  They are defined,
respectively as the smallest positive solutions to the inequalities
\begin{align}
\label{eqn:critical}
 \frac{\subgausspop_\numobs(\popgausscrit,
   \shiftfclass)}{\popgausscrit} \leq \frac{\popgausscrit}{2}, \quad
 \text{and} \quad \frac{\radcomp_\numobs(\empcrit;
   \shiftfclass)}{\empcrit} \leq \frac{\empcrit}{\bound}.
\end{align}
As long as the function class $\shiftfclass$ is convex (as assumed
here), it can be shown that both of the functions $\delta \mapsto
\frac{\subgausspop_\numobs(\delta)}{\delta}$ and $\delta \mapsto
\frac{\radcomp_\numobs(\delta)}{\delta}$ are non-increasing on the
positive real line, and so the critical radii always exist.  We refer
the reader to Chapters 13 and 14 of the book~\cite{dinosaur2019} for
more details on critical radii, and the arguments used to establish
claims of this type.

\subsubsection{Examples of statistical error}

Let us illustrate this general set-up by describing the form of the
critical radii for several different function classes.  For
simplicity, we consider classes with $\inftynorm{\values} = \bound =
1$.  Here we summarize the form of the critical radii;
see~\Cref{sec:critical-computations} for the computations that
underlie these conclusions.

\paragraph{Polynomials of degree $\polydeg$:} As our first example,
suppose that $\shiftfclass$ is the family of $\polydeg$-degree
polynomials defined on $\stateset = [-1, 1]$ uniformly bounded by
$1$. For a given $\coeffs \in \RR^{\polydeg + 1}$, define
$f_\coeffs(x) \defn \coeffs_0 + \coeffs_1 x + \ldots +
\coeffs_{\polydeg + 1} x^\polydeg$. Then we have
\begin{align*}
\shiftfclass \defn \{f_\coeffs: [-1, 1] \to \RR \text{ for some
  $\coeffs \in \RR^{\polydeg + 1}$ such that $\inftynorm{f_\coeffs}
  \leq 1$ } \}.
\end{align*}
Some straightforward calculations yield $\empcrit^2 \lesssim
\frac{\polydeg+1}{\numobs}$, and $\popgausscrit^2 \lesssim
\frac{\polydeg+1}{\numobs}$, so we can conclude
\begin{align*}
\critradius^2 \lesssim \frac{\polydeg + 1}{\numobs}.
\end{align*}

\paragraph{Radial basis kernels:} Consider $\stateset = [-1, 1]$
and the reproducing kernel Hilbert space (RKHS) $\RKHSspace$ given by
the Gaussian kernel $\kernel(x, z) = e^{-\tfrac{1}{2}(x - z)^2}$
defined on the Cartesian product space $[-1, 1] \times [-1, 1]$.
Consider the unit ball in this RKHS, given by
\begin{align*}
\shiftfclass = \{ f \in \RKHSspace \mid \RKHSnorm{f} \leq 1,
\inftynorm{f} \leq 1 \}.
\end{align*}
For this example, it can be shown that $\empcrit^2 \lesssim
\frac{\log(\numobs + 1)}{\numobs}$ and $\popgausscrit^2 \lesssim
\frac{\log(\numobs + 1)}{\numobs}$, and hence
\begin{align*}
\critradius^2 \lesssim \frac{\log(\numobs + 1)}{\numobs}.
\end{align*}
This is a simple example of a non-parametric class, since the
effective degrees of freedom ($\log (\numobs+1)$ in this case) grows
as a function of the sample size.

\paragraph{Convex Lipschitz functions:} We let $\stateset = [0, 1]$
and $\shiftfclass$ be the class of convex $1$-Lipschitz functions,
i.e.
\begin{align*}
\shiftfclass \defn \{ f :[0, 1] \to \RR \mid f(0) = 0, \text{ and $f$
  is convex and $1$-Lipschitz} \}.
\end{align*}
We can show $ \empcrit^2 \lesssim \frac{1}{n^{4/5}}$ and $\popgausscrit^2 \lesssim \frac{1}{n^{4/5}}$, giving
\begin{align*}
\critradius^2 \lesssim \frac{1}{n^{4/5}}.
\end{align*}


\section{Numerical results}
\label{sec:plots}

We illustrate the behavior of our algorithm in a mixture of idealized
and semi-realistic settings.  \Cref{sec:plots-tabular} highlights the
behavior of our algorithm in the tabular setting, where $\stateset$ is
finite, and \Cref{sec:plots-LQR} illustrates the performance in the
linear-quadratic setting, where $\stateset = \RR^d$, but the
transitions and costs have simple
representations. \Cref{sec:plots-NonLinear} demonstrates the behavior
of our algorithm for a nonlinear system with more complicated
dynamics. \Cref{sec:plots-ARCH} presents our algorithm on another nonlinear system with
dynamics, the ARCH model used in econometrics with quadratic cost.

We compare our algorithm to value iteration and fitted value
iteration. Recall that the Bellman operator $\bellmanOp$ is
$\contractPar$-contractive with respect to $\munorm{\cdot}$, so we can
iteratively apply $\bellmanOp$ and ensure convergence to the true
value function $\valuestar$; this is the value iteration algorithm.
The fitted value iteration (FVI) algorithm applies when the Bellman
update \emph{cannot} be computed $\bellmanOp(\values)$ exactly, and
must be approximated.  Given a dataset $\data = \{ (\state_i, \rsam_i,
\state_i') \}_{i=1}^\numobs$ and a function class $\fclass$, we
compute an approximation via
\begin{align*}
\widehat{\bellmanOp(\values)} \in \arg \min_{f \in \fclass} \left \{
\frac{1}{n} \sum_{i=1}^\numobs \big[ \rsam_i + \contractPar
  \values(\state_i') - f(\state_i) \big]^2 \right \}.
\end{align*}
In fitted value iteration, we iterate repeating this regression
procedure for every step (potentially with a different
dataset). Algorithm~\ref{KBB}, in each iteration, computes the Bellman
residual $\bellmanRes(\values_t) = \values_t - \bellmanOp \values_t$,
which involves computing the Bellman operator $\bellmanOp \values_t$.
Accordingly, we compare each iteration to one iteration of value
iteration and fitted value iteration, using the same regression
procedure and number of samples. In our implementations of
Algorithm~\ref{KBB}, we use $\LSTDdata_t = \REGdata_t$---that is, the
same samples used in fitting the regression procedure are re-used for
the LSTD fit. When running the algorithm, we have observed that
ensuring the first regression is very accurate results in much better
convergence. In the first step, Krylov-Bellman is estimating the
reward function $\reward$. It is likely essential to an accurate
estimate of the reward function since it corresponds to the right-hand
side of the linear operator equation $\opQ \values = \reward$.


\subsection{Tabular MRPs}
\label{sec:plots-tabular}

In a tabular MRP, the state space $\stateset$ is finite; for ease of
notation we represent it as \mbox{$\stateset = \{1, 2, \ldots, \dims
  \}$} where $\dims = \numstates$. We can then represent all
quantities of interest as either a matrix or a vector: the transition
operator $\TransOp$ can be written as a matrix $\TransMat \in
\RR^{\numstates \times \numstates}$ and the reward function $\reward$
can be identified with a vector in $\reward \in \RR^{\numstates}$ via
\begin{align*}
\reward_x \defn \reward(x) \qquad \text{and} \qquad \TransMat_{\state,
  \state'} = \TransOp(\state' \mid \state) \qquad \text{for all
  $\state \in \stateset$},
\end{align*}
where $\TransOp(\state' \mid \state)$ is the probability of transition
to $\state'$ from $\state$. Similarly we can represent a value
function $\values$ as an equivalent vector in $\RR^{\numstates}$. The
Bellman operator can then be written in the matrix form as $\bellmanOp
\values = \reward + \contractPar \TransMat \values$. In this setting
we perform the regression procedure over the function class $\fclass =
\RR^{\numstates}$, and the regression is equivalent to computing the
sample average for every state.

\begin{figure}[h]
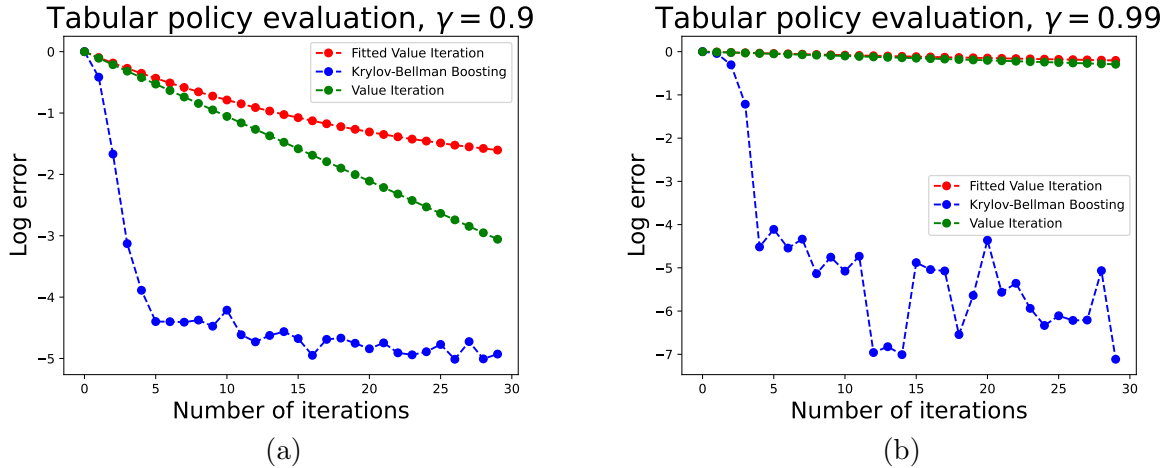

  \begin{center}
    \begin{tabular}{cc}
  \widgraph{0.5\textwidth}{\figdir/tabular_gamma0.9} &
  \widgraph{0.5\textwidth}{\figdir/tabular_gamma0.99} \\ (a) & (b)
    \end{tabular}
    \caption{Illustration of the behavior of three different
      algorithms, value iteration (red), fitted value iteration
      (green), and Krylov-Bellman Boosting (blue) on the random
      tabular MRP. We plot the log error in the $\munorm{\cdot}$-norm
      (vertical axis) against the number of iterations used
      (horizontal axis). (a) MRP with discount factor $\contractPar =
      0.9$. Value iteration exhibits the characteristic linear plot
      that reflects the $\contractPar$-contractive nature of value
      iteration. Fitted value iteration approximates value iteration,
      and performs worse as a result of being an inexact
      procedure. Krylov-Bellman Boosting performs much better than
      either of these algorithms, although we see that its convergence
      slows down drastically and becomes somewhat unstable. (b) MRP
      with discount factor $\contractPar = 0.99$. Like previously,
      value-iteration exhibits a linear plot that illustrates the
      contractivity of the Bellman operator $\bellmanOp$, albeit at a
      much slower rate. Fitted value iteration is also still slower
      than value iteration, and Krylov--Bellman Boosting is much
      faster than both of them, but exhibits much greater instability
      in this setting.}
    \label{fig:rand-tabular}
  \end{center}
\end{figure}

\paragraph{Random Transitions:}
In our first example of a tabular MRP, we create a $\numstates = 300$
dimensional MRP by generating a transition matrix $\TransMat \in
\RR^{300 \times 300}$ with i.i.d. $\text{Unif}\,(0, 1)$ entries, with
each row rescaled so as to ensure row-stochasticity. Similarly, we
generate a reward vector $\reward \in \RR^{300}$ with i.i.d
$\text{Unif}\,(0, 1)$ entries. We generate such independent samples of
such models, and set the discount parameter as $\contractPar = 0.9$ in
one case, and $\contractPar = 0.99$ in the other. The results are
plotted in Figure~\ref{fig:rand-tabular}. The plots clearly illustrate
the advantages of Algorithm~\ref{KBB} over fitted value iteration and
value iteration.  As the KBB error becomes quite small, we see the
error floor (due to statistical error in evaluating the residual)
start to dominate, as should be expected.

\begin{figure}[h]
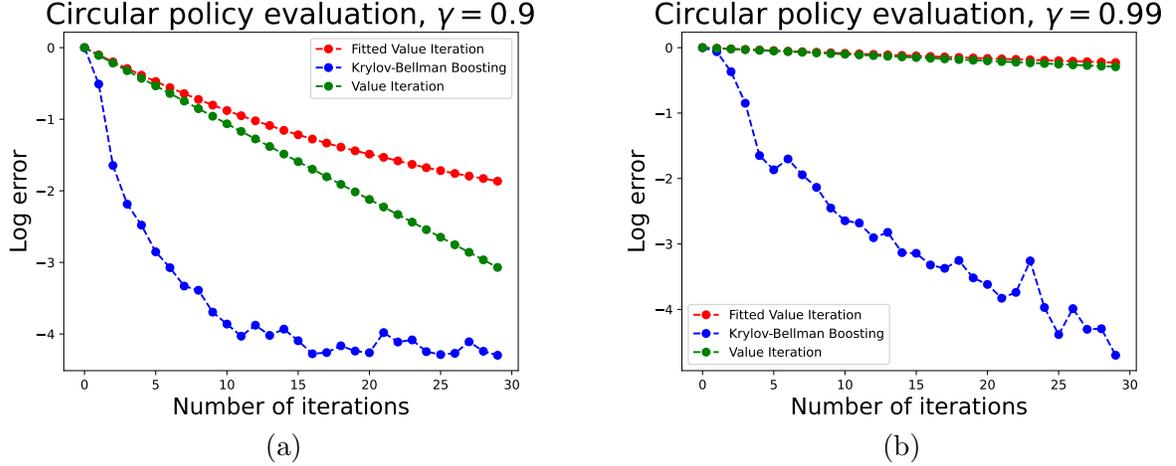

  \begin{center}
    \begin{tabular}{cc}
  \widgraph{0.5\textwidth}{\figdir/circular_gamma0.9} &
  \widgraph{0.5\textwidth}{\figdir/circular_gamma0.99} \\ (a) & (b)
    \end{tabular}
    \caption{Illustration of the behavior of three different
      algorithms, value iteration (red), fitted value iteration
      (green), and Krylov-Bellman Boosting (blue) on the random
      tabular MRP. We plot the log error in the $\munorm{\cdot}$-norm
      (vertical axis) against the number of iterations used
      (horizontal axis). The algorithms in this setting perform quite
      similarly to those in Figure~\ref{fig:rand-tabular}, although
      the Krylov--Bellman Boosting algorithm seems to be more stable
      in this setting than in the previous one. (a) Circular MRP with
      discount factor $\contractPar = 0.9$. (b) Circular MRP with
      discount factor $\contractPar = 0.99$.}
    \label{fig:circ-tabular}
  \end{center}
\end{figure}

\paragraph{Circular Random Walk:} For a more structured
model, we create a tabular MDP with $\numstates = 200$ states as
follows: indexing the states modulo 200 (i.e. state $-2$ is actually
state $198$), for a state $\state$ it has $\frac{1}{3}$ probability
of transitioning to itself, and then a $\frac{1}{6}$ probability of
transitioning to states $\state - 2, \state - 1, \state + 1$, and
$\state + 2$. The dynamics here model a random walk on the circle, and
the transition operator has a more interesting eigenstructure than the purely random
transition models from the previous setting.  As shown in
Figure~\ref{fig:circ-tabular}, the qualitative behavior of the
algorithms is similar.  The value iteration algorithms, either exact
or fitted, still exhibits the linear behavior that is indicated by the
$\contractPar$-contractive nature of the Bellman operator
$\bellmanOp$. Krylov--Bellman Boosting is again much faster than
either of the algorithms, especially in the case of $\contractPar =
0.99$.

\subsection{Linear Quadratic Regulator}
\label{sec:plots-LQR}

The linear quadratic regulator (LQR) is a canonical model in optimal
control; it plays a central role in many applications including
aeronautics, robotics and industrial process control~\cite{Bert17vol1,
  kalman1960contr}. Here we consider a particular variant of the LQR
problem known as the linear quadratic Gaussian (LQG) problem. The
state space $\stateset = \RR^{\dims}$ and action space $\actionset =
\RR^{\actdim}$ are continuous, and the transition dynamics starting
from state $\state_t \in \RR^\dims$ and taking action $\action \in
\RR^\actdim$ is given by
\begin{align*}
\state_{t+1} = \LQRtrans \state_t + \LQRact \action_t + \Noise_t,
\end{align*}
where $\LQRtrans \in \RR^{\dims \times \dims}$ is the transition
matrix, $\LQRact \in \RR^{\dims \times \actdim}$ is the action matrix,
and $\noise_t \sim \normal(0, \covmat)$. We formulate this problem in
terms of a cost function, rather than a reward function (although for
the purposes of policy evaluation it is a meaningless
distinction). Given symmetric positive semidefinite matrices
$\LQRscost \in \RR^{\dims \times \dims}$ and $\LQRacost \in
\RR^{\actdim \times \actdim}$, the cost function at each stage is
\begin{align*}
\cost(\state, \action) = \state^T \LQRscost \state + \action^T
\LQRacost \action.
\end{align*}
We restrict our attention to policies of the form $\action_t =
\LQRcontrol \state_t$ where $\LQRcontrol \in \RR^{\actdim \times
  \dims}$. We then have a Markov reward process with dynamics and
costs given by
\begin{align*}
\state_{t+1} = \big( \LQRtrans + \LQRact \LQRcontrol \big) \state_t +
\noise_t \qquad\text{and} \qquad \cost(\state_t) = \state_t^T \big(
\LQRscost + \LQRcontrol^T \LQRacost \LQRcontrol \big) \state_t.
\end{align*}
Assume that the eigenvalues of $\LQRtrans + \LQRact \LQRcontrol$ are
contained within the unit circle on the complex plane. The goal is to
solve for the value function $\valuestar$ of this MRP. The value
function can be written as
\begin{align*}
\valuestar(\state) = \state^T \lyastar \state + \frac{\contractPar}{1
  - \contractPar} \cdot \text{trace}(\lyastar \covmat),
\end{align*}
where the matrix $\lyastar \in \RR^{\dims \times \dims}$ is the
solution to the Lyapunov equation
\begin{align*}
\lyastar = \LQRscost + \LQRcontrol^T \LQRacost \LQRcontrol +
\contractPar \big(\LQRtrans + \LQRact \LQRcontrol \big)^T \lyastar
\big(\LQRtrans + \LQRact \LQRcontrol \big).
\end{align*}
Value iteration involves computing the update
\begin{align*}
\lya_{t+1} = \LQRscost + \LQRcontrol^T \LQRacost \LQRcontrol +
\contractPar \big(\LQRtrans + \LQRact \LQRcontrol \big)^T \lya_t
\big(\LQRtrans + \LQRact \LQRcontrol \big) \quad \text{and} \quad
c_{t+1} = \contractPar c_t + \contractPar \text{trace}(\lya_t
\covmat).
\end{align*}
The pair $(\lya_y, c_t)$ define the value function via  
$\values_{t}(\state) = \state^T \lya_t \state + c_t$.
When $\LQRtrans + \LQRact \LQRcontrol$ has eigenvalues contained
within the unit circle, this update is guaranteed to be a contraction.

\begin{figure}[h]
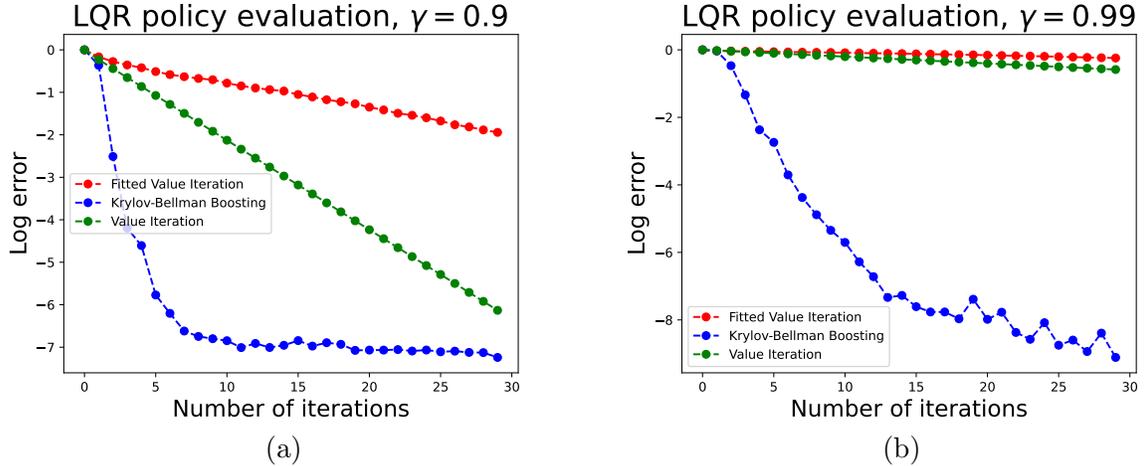

  \begin{center}
    \begin{tabular}{cc}
  \widgraph{0.5\textwidth}{\figdir/LQR_gamma0.9} &
  \widgraph{0.5\textwidth}{\figdir/LQR_gamma0.99} \\ (a) & (b)
    \end{tabular}
    \caption{Illustration of the behavior of three different
      algorithms, value iteration (red), fitted value iteration
      (green), and Krylov-Bellman Boosting (blue) on the LQR policy
      evaluation problem. We plot the log error in the
      $\munorm{\cdot}$-norm (vertical axis) against the number of
      iterations used (horizontal axis). The algorithms in this
      setting perform quite similarly to those in
      Figure~\ref{fig:rand-tabular}, although the Krylov--Bellman
      Boosting algorithm seems to be more stable in this setting than
      in the previous one. (a) Linear Quadratic Regulator with
      discount factor $\contractPar = 0.9$. (b) Linear Quadratic
      Regulator with discount factor $\contractPar = 0.99$.}
    \label{fig:LQR-plots}
  \end{center}
\end{figure}

In the numerical simulations, we choose $\dims = 5$ and $\actdim =
3$. The matrices $\LQRtrans$, $\LQRact$, and $\LQRcontrol$ are chosen
to have i.i.d.~$\text{Unif}\,(0, 1)$ entries, and then rescaled to
ensure that the eigenvalues are contained within the unit circle. The
symmetric matrices $\LQRscost$ and $\LQRacost$ are computed by
generating matrices with i.i.d.~$\text{Unif}\,(0, 1)$ random variables
and then left-multiplied by its transpose to ensure symmetry. The
regression steps in both fitted value iteration and Krylov--Bellman
Boosting use the XGBoost routine~\cite{chen2016xgboost} to perform the
fitting. Result are given in Figure~\ref{fig:LQR-plots}. As before, we
observe a geometric (or linear) convergence rate for value iteration,
consistent with the $\discount$-contractive nature of the update. We
also see approximately linear convergence of fitted value iteration,
albeit at a slower rate compared to value iteration. Krylov--Bellman
Boosting is again more rapidly convergent than these two procedures.
(We note that KBB hits an error floor due to the statistical sampling
involved, whereas exact value iteration, which is based on unrealistic
knowledge of the true dynamics, does not involve any such error.)  For
a larger discount factor $\contractPar = 0.99$, the gains afforded by
the KBB procedure become more significant.

\subsection{A non-linear example}
\label{sec:plots-NonLinear}

Consider the nonlinear system in the $3$-dimensional state variable
$\state_t = \begin{bmatrix} \state_{t,1} & \state_{t,2} & \state_{t,3}
\end{bmatrix} \in \real^3$ and control vector $\action_t \in \real^3$
given by
\begin{align*}
\begin{bmatrix} \state_{t+1, 1} - \state_{t+1, 2}^2 \\
\state_{t+1, 2} \\
\state_{t + 1, 3} - \state_{t + 1, 1}^2 \end{bmatrix} =
\LQRtrans \begin{bmatrix} \state_{t, 1} - \state_{t, 2}^2 \\
\state_{t, 2} \\
\state_{t, 3} - \state_{t,1}^2 \end{bmatrix} + \LQRact \action_t +
\noise_t
\end{align*}
with control and cost function given by
\begin{align*}
\action_t = \LQRcontrol \begin{bmatrix} \state_{t, 1} - \state_{t,
    2}^2 \\
\state_{t, 2} \\
\state_{t, 3} - \state_{t,1}^2 \end{bmatrix} , \quad \mbox{and} \quad
c_t(\state_t, \action_t) = \begin{bmatrix} \state_{t, 1} - \state_{t,
    2}^2 \\
\state_{t, 2} \\
\state_{t, 3} - \state_{t,1}^2 \end{bmatrix}^T
\LQRscost\begin{bmatrix} \state_{t, 1} - \state_{t, 2}^2 \\
\state_{t, 2} \\
\state_{t, 3} - \state_{t,1}^2 \end{bmatrix} + \action^T \LQRacost
\action_t.
\end{align*}
Here $\LQRtrans$, $\LQRact$, and $\LQRcontrol$ are all
three-dimensional matrices, with each of $\LQRscost$ and $\LQRacost$
positive semidefinite, whereas the random vector $\noise_t \in
\real^3$ follows a zero-mean multivariate Gaussian random
distribution. We choose a nonlinear control system of this form for
ease of computation: for a system of this general type, despite being
nonlinear, it is straightforward to perform exact value iteration via
a change of coordinates (see the book~\cite{Isidori95} for further
details). In the simple case where $\noise_t = 0$ and
\begin{align*}
\LQRtrans = \begin{bmatrix} 0 & 1 & 0 \\ 0 & 0 & 1 \\ 1 & 0 &
  0 \end{bmatrix} \quad \text{and} \quad \LQRact = \begin{bmatrix} 0 &
  0 & 0 \\ 1 & 0 & 0 \\ 0 & 0 & 0 \end{bmatrix},
\end{align*}
we can verify the system has the following state space
representation
\begin{align*}
\state_{t+1, 1} &= \state_{t, 2} + \left[\state_{t, 3} - \state_{t,
    1}^2 + \action_{t, 1} \right]^2, \\ \state_{t+1, 2} &=
\state_{t,3} - \state_{t, 1}^2 + \action_{t, 1}, \qquad \qquad \text{and} \\ \state_{t+1, 3} &=
\state_{t, 1} + 2\state_{t, 2} \left[ \state_{t, 3} - \state_{t, 1}^2
  + \action_{t ,1} \right]^2 + \left[\state_{t, 3} - \state_{t, 1}^2 +
  \action_{t, 1}\right]^4.
\end{align*} 
We generate the associated matrices $\LQRtrans, \LQRact, \LQRscost,
\LQRcontrol$, and $\LQRacost$ in the same manner
as~\Cref{sec:plots-LQR}.

\begin{figure}[h]
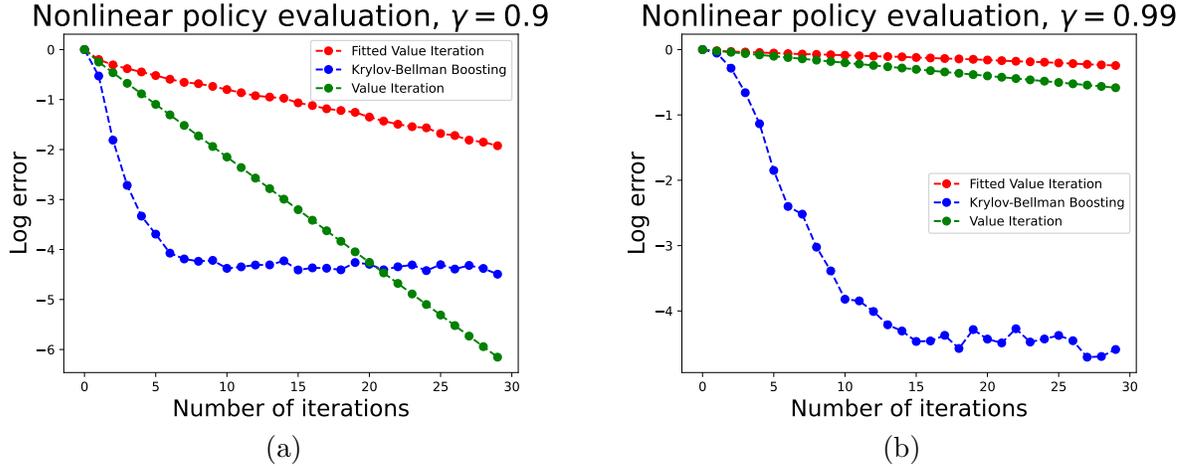

  \begin{center}
    \begin{tabular}{cc}
  \widgraph{0.5\textwidth}{\figdir/nonlinear_gamma0.9} &
  \widgraph{0.5\textwidth}{\figdir/nonlinear_gamma0.99} \\ (a) & (b)
    \end{tabular}
    \caption{Illustration of the behavior of three different
      algorithms, value iteration (red), fitted value iteration
      (green), and Krylov-Bellman Boosting (blue) on the nonlinear
      policy evaluation problem. We plot the log error in the
      $\munorm{\cdot}$-norm (vertical axis) against the number of
      iterations used (horizontal axis). Algorithm~\ref{KBB} is still
      much faster than fitted-value iteration, but in the case with
      $\contractPar = 0.9$ the Krylov--Bellman Boosting is faster than
      value iteration initially but because of the noise, value
      iteration catches up and overtakes it. (a) Discount factor
      $\contractPar = 0.9$. (b) Discount factor $\contractPar =
      0.99$.}
    \label{fig:nonlin-plots}
  \end{center}
\end{figure}

The results of the algorithm are plotted in
Figure~\ref{fig:nonlin-plots}. We still observe the linear behavior of
value iteration in both settings, however this algorithm requires us
to know the underlying nonlinear system exactly as well as its transition
dynamics. In general, Algorithm~\ref{KBB} performs much better than
fitted value iteration. When $\contractPar = 0.9$, initially the
algorithm is much faster than value iteration, but eventually it
catches up and overtakes it due to the noise in the algorithm. In the
setting of $\contractPar = 0.99$, we see that Krylov--Bellman Boosting
is much faster than either value iteration or fitted value
iteration. In this setting, we are running Krylov--Bellman
Boosting and fitted value iteration on the state space given by $\state_t$, which has nonlinear
transition dynamics and cost functions.

\subsection{ARCH}
\label{sec:plots-ARCH}

For the final example, we consider the autoregressive conditional
heteroskedasticity model (ARCH)~\cite{engle1982ARCH,
  BOLLERSLEV1986GARCH}.  The state vectors $\State_t$ evolve according
to the non-linear dynamics
\begin{align*}
\State_{t+1} = \LQRtrans \State_t + \sqrt{q + \State_t \LQRscost
  \State_t } \cdot \Noise_t
\end{align*}
where $q \geq 0$ is a scalar, the matrix $\LQRscost \in \RR^{d \times
  d}$ is positive semidefinite, and $\Noise_t \sim \normal(0,
\covmat)$ is system noise. We turn this into a Markov reward process
by assigning a cost function $\cost(\state) = \state^T \LQRacost
\state$ and adding discounting. The value function $\valuestar$
is the solution to the Bellman relation
\begin{align*}
\values(\state) = \bellmanOp(\values)(\state) \defn \state^T \LQRacost
\state + \contractPar \EE_\Noise \left[ \values(\LQRtrans \state +
  \sqrt{q + \state^T \LQRscost \state} \cdot \Noise) \right].
\end{align*}
Under some regularity conditions on the triple $(q, \LQRtrans,
\LQRacost)$, the Bellman operator $\bellmanOp$ is
$\contractPar$-contractive under the norm $\statnorm{\cdot}$.
Consequently, we can solve for the value function via value iteration
in the same manner as previous settings. A straightforward calculation
yields
\begin{align*}
\valuestar(\state) = \state^T \lyastar \state + \frac{\contractPar
  q}{1 - \contractPar} \cdot \text{trace}(\lyastar \covmat)
\end{align*}
where the matrix $\lyastar \in \RR^{d \times d}$ solves the linear
system
\begin{align*}
\lyastar = \LQRacost + \contractPar \left\{ \LQRtrans^T \lyastar
\LQRtrans + \LQRscost \cdot \text{trace}(\lyastar \covmat) \right\}.
\end{align*}

\begin{figure}[h]
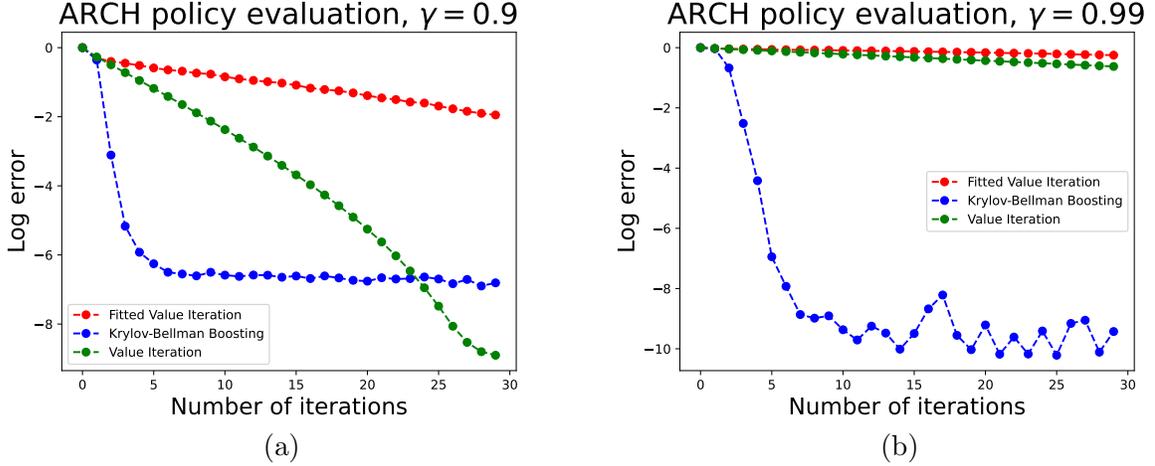

  \begin{center}
    \begin{tabular}{cc}
  \widgraph{0.5\textwidth}{\figdir/ARCH_gamma0.9} &
  \widgraph{0.5\textwidth}{\figdir/ARCH_gamma0.99} \\ (a) & (b)
    \end{tabular}
    \caption{Illustration of the behavior of three different
      algorithms, value iteration (red), fitted value iteration
      (green), and Krylov Bellman Boosting (blue) on the ARCH policy
      evaluation problem. We plot the log error in the
      $\munorm{\cdot}$-norm (vertical axis) against the number of
      iterations used (horizontal axis). Algorithm~\ref{KBB} is still
      much faster than fitted-value iteration, but in the case with
      $\contractPar = 0.9$ the Krylov--Bellman Boosting is faster than
      value iteration initially but because of the noise, value
      iteration catches up and overtakes it. (a) ARCH with discount
      factor $\contractPar = 0.9$. (b) ARCH with discount factor
      $\contractPar = 0.99$}
    \label{fig:ARCH-plots}
  \end{center}
\end{figure}

In our numerical simulations we choose $d = 5$, and $q = 0.5$. The
matrix $\LQRtrans$ is chosen to have i.i.d.~Unif(0, 1) entries, and
then rescaled to ensure that the system is stable. The matrix
$\LQRscost$ is computed by generating a matrix with
i.i.d.~$\mbox{Unif}(0, 1)$ random variables, and then left-multiplying
by its transpose to ensure symmetry.  The results are plotted in
Figure~\ref{fig:ARCH-plots}. The behavior in this setting seems to be
the same as what we observed in the past few settings. Initially,
Krylov--Bellman Boosting is much faster than both fitted value
iteration and value iteration. In the setting of $\contractPar = 0.9$,
Krylov Bellman Boosting hits an error floor relatively early, and it
takes value iteration and fitted value iteration quite a while to
catch up. For $\contractPar = 0.99$, we observe a similar behavior,
but both value iteration and fitted value iteration take significantly
longer to match Krylov--Bellman Boosting in accuracy.

\section{Proofs}
\label{sec:proofs}

We now turn to the proofs of our claims.  We begin by
proving~\Cref{thm:main} in~\Cref{SecThmMainProof}. This proof relies
on a result of possible interest---stated as~\Cref{props:one-step} and
proved in~\Cref{sec:prop-proof}---that controls the one-step of the
algorithm when the LSTD calculations are exact. \Cref{sec:regthm-proof} contains
the proof of~\ref{thm:regression}, and~\Cref{sec:reg-lem-proofs} the lemmas required.


\subsection{Proof of~\Cref{thm:main}}
\label{SecThmMainProof}

Our proof of this result consists of two parts:
\begin{enumerate}
\item[(a)] First, we analyze an idealized version of the algorithm in
  which there is no error in the LSTD calculation.  We summarize our
  conclusions in \Cref{props:one-step} stated below.
\item[(b)] Second, we leverage our understanding of the ``exact-LSTD''
  setting in order to analyze the general algorithm.  Our strategy is
  to reduce the general update to a perturbed form of the ``exact-LSTD''
  update.
\end{enumerate}

\newcommand{\lstdcurrent}{\smallsuper{V}{\lstd}}
\newcommand{\lstdnext}{\smallsuper{V}{\lstd}_{+}}

We begin by bounding the behavior of an update that is not actually
used in the algorithm itself, but is a useful auxiliary quantity.  In
particular, suppose that we are given an LSTD solution $\lstdcurrent$,
computed from a subspace $\subspace$ with spectral parameters
$(\mineig, \maxeig)$, as previously defined~\eqref{eqn:ortho-eig}.
For a stepsize $\update > 0$, define the update
\begin{align*}
  \OneStep_\update(\lstdcurrent) & = \lstdcurrent - \update \big \{
  \bellmanRes(\lstdcurrent) + \REGerr \big \},
\end{align*}
where $\REGerr$ represents noise in the computation of the Bellman
residual. We let $\bias = \EE[\REGerr]$ and $\REGvar =
\EE\Qnorm{\REGerr - \bias}^2$ denote the bias and variance,
respectively, of this noise. Recall that by our regression-noise
assumption~\ref{EqnREGvar}, the mean-squared error satisfies the upper
bound \mbox{$\Qnorm{\bias}^2 + \REGvar \leq \REGMSE^2$}---we say that
the regression procedure is $\REGMSE$-accurate for short.

\begin{proposition}
\label{props:one-step}
Given a $\REGMSE$-accurate regression procedure and a stepsize
\mbox{$\update \in [0, \frac{1}{2}]$,} for any LSTD solution
$\lstdcurrent$, the update $\OneStep_\update(\lstdcurrent)$ satisfies
the bound
\begin{subequations}
  \begin{align}
\label{eqn:prop-initial}    
 \Exs \Qnorm{\OneStep_\update(\lstdcurrent) - \valuestar}^2 & \leq
 \big(1 - \tfrac{3 \update \mineig}{2} + 2 \update^2 \maxeig \big)
 \Qnorm{\valuesLSTD - \valuestar}^2 + \big(\tfrac{8 \update}{\mineig}
 + \update^2 \big) \Qnorm{\bias}^2 + \update^2 \cdot \REGvar,
\end{align}
 where $\EE$ denotes expectation over the regression noise $\REGerr$.
 In particular, for the choice $\update = \frac{\mineig}{2\maxeig}$,
 we have
\begin{align}
\label{eqn:prop-guarantee}
\Exs \Qnorm{\OneStep_\update(\lstdcurrent) - \valuestar}^2 & \leq
\big(1 - \tfrac{\mineig^2}{4 \maxeig} \big)\Qnorm{\lstdcurrent -
  \valuestar}^2 + \tfrac{5}{\maxeig} \Qnorm{\bias}^2 +
\tfrac{1}{2\maxeig} \cdot \REGvar.
\end{align}
\end{subequations}
\end{proposition}
\noindent See~\Cref{sec:prop-proof} for the proof of this
proposition. \\

In addition we require another auxiliary result, which characterizes
the LSTD solution as a projection under the $\opQ$-norm:
\begin{lemma}
\label{lems:lstd-minimizer}
For a reversible Markov chain, the LSTD estimate $\valuesLSTD$ is the
projection of $\valuestar$ onto $\subspace$ under the
$\Qnorm{\cdot}$-norm---viz.
\begin{align}
\valuesLSTD & = \arg \min_{\values \in \subspace} \Qnorm{\values -
  \valuestar}^2.
\end{align}
\end{lemma}
\noindent See~\Cref{SecProofLSTDMin} for the proof. \\

\medskip

Equipped with these two auxiliary results, we are now ready to
prove~\Cref{thm:main}, which applies to the algorithm that includes
errors in both the regression and LSTD phase.  Focusing on the update
from $\values_t$ to $\values_{t+1}$, we need to bound the error
$\Qnorm{\values_{t+1} - \valuestar}$ in terms of the error
$\Qnorm{\values_t - \valuestar}$.  In order to do so, our analysis
also involves the exact LSTD solutions $\valuesLSTD_{t+1}$ and
$\valuesLSTD_{t}$; let us emphasize that these quantities are
\emph{not} actually computed in the algorithm itself.

Condition (a) in the definition~\ref{EqnLSTDerr} of LSTD accuracy
allows us to relate $\values_{t+1}$ to $\valuesLSTD_{t+1}$ via the
inequality
\begin{subequations}
\begin{align}
  \Qnorm{\values_{t+1} - \valuestar}^2 & \leq \Qnorm{\valuesLSTD_{t+1}
    - \valuestar}^2 + (\LSTDerr_t)^2.
\end{align}
Next, we observe that for any scalar $\update$, the function
$\valuesLSTD_{t} - \update\big( \bellmanRes(\values_t) +
\REGerr_{t+1}\big)$ is an element of $\subspace_{t+1}$, since
$\basis_{t+1} = \bellmanRes(\values_t) + \REGerr_{t+1}$ is the basis
function added at round $t$.  Consequently,
applying~\Cref{lems:lstd-minimizer} with $\valuesLSTD_{t+1}$ and the
subspace $\subspace_{t+1}$ guarantees that
\begin{align}
\Qnorm{\valuesLSTD_{t+1} - \valuestar}^2 & \leq \Qnorm{\valuesLSTD_{t}
  - \update\big( \bellmanRes(\values_t) + \REGerr_{t+1} \big) -
  \valuestar}^2 .
\end{align}
\end{subequations}
Now observe that, by definition of the one-step update
$\OneStep_\update$, we have the equivalence
\begin{align*}
\valuesLSTD_{t} - \update\big( \bellmanRes(\values_t) + \REGerr_{t+1}
\big) & = \OneStep_\update(\valuesLSTD_t) + \update \big \{
\bellmanRes(\valuesLSTD_t) - \bellmanRes(\values_t) \big \}.
\end{align*}
Using this decomposition and applying the Fenchel-Young inequality
with a parameter $\tau > 0$ to be chosen, we find that
\begin{align*}
  \Qnorm{\values_{t+1} - \valuestar}^2 & \leq (1 + \tau)
  \Qnorm{\OneStep_\update(\valuesLSTD_t) - \valuestar}^2 + \update^2 \big(1 +
  \tfrac{1}{\tau} \big) \Qnorm{\bellmanRes(\values_t) -
    \bellmanRes(\valuesLSTD_t)}^2 + (\LSTDerr_t)^2 \\
& \stackrel{(i)}{\leq} (1 + \tau)
  \Qnorm{\OneStep_\update(\valuesLSTD_t) - \valuestar}^2 + \big(2 +
  \tfrac{1}{\tau} \big) (\LSTDerr)^2,
\end{align*}
where step (i) follows inequality (b) in the definition of
$\LSTDerr$-accuracy for the LSTD computation as well as the fact that $\update \leq \frac{1}{2}$.

We can apply~\Cref{props:one-step} with the choice $\update =
\frac{\mineig_t}{2\maxeig_t}$ to upper bound the first term, thereby
obtaining
\begin{align*}
\EE_{t+1}\big[\Qnorm{\OneStep_\update(\valuesLSTD_t) - \valuestar}^2
  \big] & \leq \kappa_t \Qnorm{\valuesLSTD_t - \valuestar}^2 +
\tfrac{5}{\maxeig_t} \Qnorm{\bias_{t+1}}^2 + \tfrac{1}{2 \maxeig_t}
\cdot \REGvar_t.
\end{align*}
where we have introduced the shorthand $\kappa_t \defn 1 -
\frac{\mineig_t^2}{4 \maxeig_t} \in (0,1)$. Putting together the
pieces, we have
\begin{align*}
\EE_{t+1} \Qnorm{\values_{t+1} - \valuestar}^2 &\leq \left(1 +
\tau\right) \kappa_t \Qnorm{\valuesLSTD_t - \valuestar}^2 + \left(1 +
\tau \right) \left( \tfrac{5}{\maxeig_t} \Qnorm{\bias_{t+1}}^2 +
\tfrac{1}{2\maxeig_t} \cdot \REGvar_t \right) + \left( 2 +
\tfrac{1}{\tau} \right) (\LSTDerr)^2.
\end{align*}
Setting the Fenchel-Young parameter as $\tau = \frac{1 -
  \kappa_t}{2\kappa_t}$ yields
\begin{align*}
(1 + \tau) \, \kappa_t = \frac{1 + \kappa_t}{2} = 1 -
  \frac{\mineig_t^2}{8\maxeig_t}.
\end{align*}
We also have $ \tau = \frac{1 - \kappa_t}{2 \kappa_t} =
\frac{\mineig_t^2}{8 \maxeig_t - 2 \mineig_t^2} \leq \frac{1}{2},$
using the fact that $\mineig_t^2 \leq 2 \maxeig_t$ in the
inequality. Moreover, we observe that
\begin{align*}
2 + \frac{1}{\tau} = 2 + \frac{2\kappa_t}{1 - \kappa_t} = \frac{2}{1 -
  \kappa_t} = \frac{8 \maxeig_t}{\mineig_t^2}.
\end{align*}
Putting together the pieces, we conclude
\begin{align}
\label{EqnMina}
\EE_{t+1} \Qnorm{\values_{t+1} - \valuestar}^2 & \leq \big(1 -
\frac{\mineig_t^2}{8 \maxeig_t} \big) \Qnorm{\valuesLSTD_t -
  \valuestar}^2 + \frac{10}{\maxeig_t} \cdot \Qnorm{\bias_{t+1}}^2 +
\frac{1}{\maxeig_t} \cdot \REGvar_t + \frac{8\maxeig_t}{\mineig_t^2}
\cdot (\LSTDerr)^2.
\end{align}

Finally, since both $\valuesLSTD_t$ and $\values_t$ belong to the
subspace $\subspace_t$, Lemma~\ref{lems:lstd-minimizer} guarantees
that
\begin{align*}
\Qnorm{\valuesLSTD_t - \valuestar}^2 \leq \Qnorm{\values_t -
  \valuestar}^2.
\end{align*}
Substituting this upper bound into the inequality~\eqref{EqnMina} and
using the fact that $ \Qnorm{\bias_{t+1}}^2 + \REGvar_t \leq
\REGMSE_t^2$ yields the claim.


\subsection{Proof of Proposition~\ref{props:one-step}}
\label{sec:prop-proof}

Let $\valuesLSTD$ be an LSTD solution defined by the subspace
$\subspace$, and let $(\mineig, \maxeig)$ be the associated spectral
quantities.  Recall that our goal is to control the behavior of the
operator $\OneStep_\update(\valuesLSTD) = \valuesLSTD - \update \big
\{\bellmanRes(\valuesLSTD) + \REGerr \big \}$, where $\REGerr$ is the
noise in the evaluation of the Bellman residual.

We first claim that the following decomposition holds
\begin{align}
\label{EqnEric}  
  \EE \Qnorm{\OneStep_\update(\valuesLSTD) - \valuestar}^2 & = \Term_1
  + \Term_2 + \update^2 \Qnorm{\bias}^2 + \update^2 \REGvar,
\end{align}
where the two terms are defined as
\begin{subequations}
\label{EqnEricTerms}  
\begin{align}
\Term_1 & \defn \Qnorm{\valuesLSTD - \valuestar}^2 - 2 \update
\munorm{\bellmanRes(\valuesLSTD)}^2 + \update^2
\Qnorm{\bellmanRes(\valuesLSTD)}^2, \quad \mbox{and} \\
\Term_2 & \defn 2 \update^2 \qinprod{\valuesLSTD - \valuestar}{\opQ
  \bias} - 2 \update \qinprod{\valuesLSTD - \valuestar}{\bias}.
\end{align}
\end{subequations}

In order to prove the claim~\eqref{EqnEric}, we begin by observing
that for any stepsize $\update > 0$, expanding the square yields
\begin{align*}
  \Qnorm{\OneStep_\update(\valuesLSTD) - \valuestar}^2 & =
  \Qnorm{\valuesLSTD - \valuestar}^2 - 2 \update \qinprod{\valuesLSTD
    - \valuestar}{\bellmanRes(\valuesLSTD) + \REGerr} + \update^2
  \Qnorm{\bellmanRes(\valuesLSTD) + \REGerr}^2,
\end{align*}
Taking expectation over the regression noise, we find that
\begin{align*}
\EE \qinprod{\valuesLSTD - \valuestar}{\bellmanRes(\valuesLSTD) +
  \REGerr} & = \qinprod{\valuesLSTD -
  \valuestar}{\bellmanRes(\valuesLSTD) + \bias}, \quad \mbox{and} \\
\EE \Qnorm{\bellmanRes(\valuesLSTD) + \REGerr}^2 & =
\Qnorm{\bellmanRes(\valuesLSTD) + \bias}^2 + \REGvar,
\end{align*}
where the second equation follows from a bias-variance
decomposition. Thus we have
\begin{align*}
\EE \Qnorm{\OneStep_\update(\valuesLSTD) - \valuestar} & =
\Qnorm{\valuesLSTD - \valuestar}^2 -2\update \, \qinprod{\valuesLSTD -
  \valuestar}{\bellmanRes(\valuesLSTD) + \bias} \\
& \qquad + \update^2 \Qnorm{\bellmanRes(\valuesLSTD) + \bias}^2 +
\update^2 \REGvar.
\end{align*}
We have
\begin{align}
\qinprod{\valuesLSTD - \valuestar}{\bellmanRes(\valuesLSTD) + b} &=
\qinprod{\valuesLSTD - \valuestar}{\bellmanRes (\valuesLSTD)} +
\qinprod{\valuesLSTD - \valuestar}{\bias} \notag \\
\label{eqn:inner-prod-term}
&= \munorm{\bellmanRes(\valuesLSTD)}^2 + \qinprod{\valuesLSTD -
  \valuestar}{ \bias},
\end{align}
and
\begin{align}
\Qnorm{\bellmanRes(\valuesLSTD) + \bias}^2 &=
\Qnorm{\bellmanRes(\valuesLSTD)}^2 + 2
\qinprod{\bellmanRes(\valuesLSTD)}{ \bias} + \Qnorm{\bias}^2 \notag \\
\label{eqn:bias-prod-term}
&= \Qnorm{\bellmanRes(\valuesLSTD)}^2 + 2 \qinprod{\valuesLSTD -
  \valuestar}{ \opQ \bias} + \Qnorm{\bias}^2,
\end{align}
using the self-adjointness of
$\opQ$. Combining~\cref{eqn:inner-prod-term,eqn:bias-prod-term} yields
\begin{align*}
\EE \Qnorm{\OneStep_\update(\valuesLSTD) - \valuestar}^2 & =
\Qnorm{\valuesLSTD - \valuestar}^2 - 2 \update
\munorm{\bellmanRes(\valuesLSTD)}^2 + \update^2
\Qnorm{\bellmanRes(\valuesLSTD)}^2 \\ & \qquad + 2 \update^2
\qinprod{\valuesLSTD - \valuestar}{\opQ \bias} - 2 \update \,
\qinprod{\valuesLSTD - \valuestar}{\bias} + \update^2 (\Qnorm{\bias}^2
+ \REGvar) \\ &= T_1 + T_2 + \update^2 (\Qnorm{\bias}^2 + \REGvar),
\end{align*}
which establishes the claimed decomposition~\eqref{EqnEric}. \\

Our next step is to bound the terms $\Term_1$ and $\Term_2$ that were
previously defined~\eqref{EqnEricTerms}.

\paragraph{Bounding $\Term_1$:}
It suffices to show that the Bellman residual
$\bellmanRes(\valuesLSTD)$ satisfies the bounds
\begin{align}
\label{eqn:norm-conversion}
\Qnorm{\bellmanRes(\valuesLSTD)}^2 \leq 2 \maxeig \Qnorm{\valuesLSTD -
  \valuestar}^2, \quad \mbox{and} \quad
\munorm{\bellmanRes(\valuesLSTD)}^2 &\geq \mineig \Qnorm{\valuesLSTD -
  \valuestar}^2.
\end{align}
Indeed, by combining these two inequalities, we find that
\begin{align*}
\Term_1 \leq \left(1 - 2\update \mineig + 2 \update^2 \maxeig \right)
\Qnorm{\valuesLSTD - \valuestar}^2,
\end{align*}

In order to prove the claims~\eqref{eqn:norm-conversion}, we first
observe that the lower bound follows immediately from the definition
of $\mineig$. As for the upper bound, we have, letting $z \defn \opQ(\valuesLSTD - \valuestar)$,
\begin{align*}
\Qnorm{z}^2 = \statinprod{\opQ^{-1/2} z}{\opQ^2 \opQ^{-1/2} z} &
\stackrel{(i)}{\leq} 2 \, \statinprod{\opQ^{-1/2} z}{\opQ \opQ^{-1/2}
  z} \\
& = 2 \, \statnorm{z}^2 \\
& \leq 2 \maxeig \Qnorm{\valuesLSTD - \valuestar}^2,
\end{align*}
where step (i) follows from the inequality $\opQ^2 \preceq 2\opQ$.

\paragraph{Bounding $\Term_2$:} We apply the  Fenchel-Young inequality
with a parameter $\tau > 0$ to be chosen, thereby obtaining the bound
\begin{align*}
|\Term_2| = 2 \update \big| \qinprod{\valuesLSTD - \valuestar}{
  (\update \opQ - \IdOp) \bias} \big| & \leq 2 \update \big(
\frac{\tau}{2} \Qnorm{\valuesLSTD - \valuestar} + \frac{1}{2\tau}
\Qnorm{(\update \opQ - \IdOp) \bias}^2 \big) \\
& = \update \tau \Qnorm{\valuesLSTD - \valuestar}^2 +
\frac{\update}{\tau} \Qnorm{(\update \opQ - \IdOp) \bias}^2.
\end{align*}
In order to bound the second term on the right-hand side of this
inequality, we again apply the Fenchel-Young inequality so as to
obtain
\begin{align*}
\Qnorm{(\update \opQ - \IdOp) \bias}^2 \leq 2\update^2 \Qnorm{\opQ
  \bias}^2 + 2 \Qnorm{\bias}^2.
\end{align*}
Since $\opQ^3 \preceq 2 \opQ^2 \preceq 4 \opQ$, we have $\Qnorm{\opQ
  \bias}^2 = \statinprod{\bias}{\opQ^3 \bias} \leq 4 \,
\statinprod{\bias}{\opQ \bias} = 4 \Qnorm{\bias}^2$, whence
\begin{align*}
\Term_2 \leq \update \tau \Qnorm{\valuesLSTD - \valuestar}^2 +
\frac{2\update}{\tau} \big(4 \update^2 + 1\big) \Qnorm{\bias}^2.
\end{align*}

We now set the Fenchel-Young parameter as $\tau = \frac{\mineig}{2}$.
With this choice, and recalling that $\update \in (0, \tfrac{1}{2}]$,
  we find that
\begin{align*}
\Term_2 \leq \frac{\update \mineig}{2} \Qnorm{\valuesLSTD -
  \valuestar} + \frac{8\update}{\mineig} \Qnorm{\bias}^2.
\end{align*}
Putting together the pieces, we conclude that
\begin{align*}
\EE \Qnorm{\OneStep_\update(\valuesLSTD)- \valuestar}^2 \leq \big(1 -
\frac{3 \update \mineig}{2} + 2 \update^2 \maxeig \big)
\Qnorm{\valuesLSTD - \valuestar}^2 + \big(\frac{8 \update}{\mineig} +
\update^2 \big) \Qnorm{\bias}^2 + \update^2 \cdot \REGvar,
\end{align*}
as claimed in~\cref{eqn:prop-initial}.

Finally, setting $\update = \frac{\mineig}{2\maxeig}$ and using the
inequality $\mineig^2 \leq 2 \maxeig$ yields the
claim~\eqref{eqn:prop-guarantee}.


\subsection{Proof of Lemma~\ref{lems:lstd-minimizer}}
\label{SecProofLSTDMin}
Recall that $\valuesLSTD$ is the LSTD solution defined by the subspace
$\subspace$. For any $\values \in \subspace$, we have
\begin{align*}
\Qnorm{\valuesLSTD- \values - \valuestar}^2 &= \Qnorm{\valuesLSTD -
  \valuestar}^2 + 2 \, \qinprod{\values}{\valuesLSTD - \valuestar} +
\Qnorm{\values}^2 \\
& = \Qnorm{\valuesLSTD - \valuestar}^2 + 2 \,
\statinprod{\values}{\opQ \big(\valuesLSTD - \valuestar \big)} +
\Qnorm{\values}^2 \\
& \stackrel{(i)}{=} \Qnorm{\valuesLSTD - \valuestar}^2 +
\Qnorm{\values}^2,
\end{align*}
where the equality (i) follows from the inclusion $\opQ(\valuesLSTD -
\valuestar) \in \subspace^\perp$, as guaranteed by definition of the
LSTD fixed point~\eqref{eqn:lstd-ortho}.  Consequently, we see that this quadratic form is
minimized by setting $\values = 0$, yielding the claim.



\subsection{Proof of~\Cref{thm:regression}}
\label{sec:regthm-proof}

Key to this proof is the notion of the \textit{empirical local
  sub-Gaussian complexity} of $\shiftfclass$, defined at scale $\delta
> 0$ as
 \begin{align*}
\subgausscomp_\numobs(\delta; \shiftfclass,
\{\state_i\}_{i=1}^\numobs) \defn \EE \left[ \sup_{g \, \in
    \shiftfclass, \, \empnorm{g} \leq \delta} \left| \frac{1}{\numobs}
  \sum_{i=1}^\numobs \noise_i g(\state_i) \right| \bigm| \{ \state_i
  \}_{i=1}^\numobs \right].
 \end{align*}
 Here the samples $\{\state_i\}_{i=1}^\numobs$ correspond to those in
 regression data set $\REGdata$, upon which we are conditioning.  For
 convenience, we drop the dependence on the observation data
 $\{\state_i\}_{i=1}^\numobs$ and directly write
 $\subgausscomp_\numobs(\delta; \shiftfclass)$. We use the shorthand
 $\EE_\noise \defn \EE[\cdot \mid \{\state_i\}_{i=1}^\numobs]$ in the
 paper occasionally for convenience.

 Relative to the population-level sub-Gaussian complexity defined
 previously~\eqref{eqn:subgausspop-def} (in which we take expectations
 over state variables), the complexity function $\subgausscomp$ treats
 the data $\{\state_i\}_{i=1}^\numobs$ as fixed where $\subgausspop$
 does not.  In analogy to the critical radii~\eqref{eqn:critical}
 defined at the population level, we also define a critical radius at
 using $\subgausscomp$: more specifically, let $\subgausscrit > 0$ be
 the smallest positive solution to the inequality
 $\frac{\subgausscomp(\delta; \shiftfclass)}{\delta} \leq
 \frac{\delta}{2}$.  Finally, recall that the dataset $\{ \state_i
 \}_{i=1}^{\numobs}$ defines the empirical norm $\empnorm{f}^2 \defn
 \frac{1}{\numobs} \sum_{i=1}^\numobs f^2(\state_i)$.

With this set-up, our proof is broken into three main steps:
\begin{enumerate}
\item[(i)] We control the empirical error $\empnorm{\fhat - \ftil}^2$
  in terms of empirical sub-Gaussian critical radius $\subgausscrit$
  and the approximation error $\empnorm{\ftil - \fstar}^2$, as stated
  in~\Cref{lems:subgauss-local}.
\item[(ii)] We establish that $\subgausscrit$ is controlled by the
  population critical radii $\popgausscrit$ and $\empcrit$, given
  by~\Cref{lems:gausscrit-expectation}.
\item[(iii)] We use the critical radius $\empcrit$ to establish
  uniform control and connect $\empnorm{\cdot}$ with
  $\statnorm{\cdot}$ (cf.~\Cref{lems:rad-local}).
\end{enumerate}

We begin by stating several technical lemmas. Recall that $\empcrit >
0$ and $\popgausscrit > 0$ satisfies the critical inequalities $\radcomp(\empcrit; \shiftfclass)
\leq \frac{\empcrit^2}{\bound}$ and $\subgausspop(\popgausscrit; \shiftfclass) \leq \frac{\popgausscrit^2}{2}$, respectively. The statement of this lemma treats
the observations $\{ \state_i \}_{i=1}^\numobs$ as fixed, and provides
control of $\empnorm{\fhat - \ftil}$ over the noise in $\noise_i$
conditional on $\{ \state_i \}_{i=1}^{\numobs}$.
\begin{lemma}
\label{lems:subgauss-local}
For any $t \geq \subgausscrit$, we have
\begin{align*}
\empnorm{\fhat - \ftil}^2 \leq c_1 t \subgausscrit + c_2
\empnorm{\ftil - \fstar}^2
\end{align*}
with probability exceeding $1 - e^{-c_3 n \cdot \frac{t
    \subgausscrit}{\inftynorm{\values}^2}}$.
\end{lemma}

\noindent
See~\Cref{sec:proof-subgauss} for the proof.

In~\Cref{lems:subgauss-local}, we have a guarantee in terms of the
empirical critical radius $\subgausscrit$, which is a random variable
dependent on $\{ \state_i \}_{i=1}^\numobs$ and $\shiftfclass$. The
following lemma controls the expectation of $\subgausscrit^2$.
\begin{lemma}
\label{lems:gausscrit-expectation}
The empirical sub-Gaussian complexity $\subgausscrit$ has expectation
bounded as
\begin{align*}
\EE \left[\subgausscrit^2\right] \leq c_1 \popgausscrit^2 + c_2 \,
\empcrit^2 + \frac{c_3}{\numobs} \left( \bound^2 +
\inftynorm{\values}^2 \right)
\end{align*}
for universal constants $c_1, c_2, c_3$.
\end{lemma}
\noindent See~\Cref{sec:proof-gausscrit-expectation} for a proof of
this lemma.

The next lemma allows us to convert our empirical norm guarantee to
one involving the stationary norm $\munorm{\cdot}$. It provides
control on $\empnorm{f - \ftil}$ in terms of $\statnorm{f - \ftil}$
uniformly for all $f \in \shiftfclass$; we require a uniform version
since $\fhat$ is a random quantity.
\begin{lemma}
\label{lems:rad-local}
For any $t \geq \empcrit$, we have
\begin{align}
\left| \empnorm{f - \ftil}^2 - \munorm{f - \ftil}^2 \right| \leq
\frac{1}{2} \munorm{f - \ftil}^2 + \frac{t^2}{2}\, \qquad \text{for
  all $f \in \shiftfclass$}
\end{align}
with probability exceeding $1 - c_1 e^{-c_2 \frac{\numobs
    t^2}{\bound^2}}$.
\end{lemma}
\noindent See Theorem 14.1 in the reference~\cite{dinosaur2019} for
the proof.

\medskip

Note for an arbitrary real-valued random variable $Z$, the
inequality $Z \leq \max\{Z, 0\}$ allows us to write $\EE[Z] \leq
\EE[\max\{Z, 0\}] = \int_0^\infty \PP(Z \geq u) \, du$, so we can
upper bound $\Exs[Z]$ by integrating an upper tail bound.

Let us apply this line of reasoning to the random variable
\begin{align*}
  Z \defn \big| \empnorm{\fhat - \ftil}^2 - \munorm{\fhat - \ftil}^2
  \big| - \frac{1}{2} \munorm{\fhat - \ftil}^2,
\end{align*}
using the tail bound established in~\Cref{lems:rad-local}. With a
change of variables, we have
\begin{align}
\begin{split}
\label{eqn:emp-control-expectation}
\EE[Z] \leq \int_0^\infty \PP(Z \geq u) \, du &\leq \frac{1}{2}
\empcrit^2 + \int_{\empcrit^2/2}^\infty \PP(Z \geq u) \, du \\ &\leq
\frac{1}{2} \empcrit^2 + \int_0^\infty c_1 e^{-2c_2 \frac{\numobs
    u}{b^2}} \, du \\ &= \frac{1}{2} \empcrit^2 + \frac{c_1}{2c_2}
\cdot \frac{b^2}{\numobs}.
\end{split}
\end{align}
Additionally, observe that~\Cref{lems:subgauss-local} treats $\{
\state_i \}_{i=1}^\numobs$ as fixed and the high probability is over
the randomness with respect to $\{ \noise_i \}$ conditional on $\{
\state_i \}$. Thus, integrating the tail bound again yields
\begin{align*}
\EE_\noise \left[ \empnorm{\fhat - \ftil}^2 \right] \leq c' \Big \{
\subgausscrit^2 + \frac{\inftynorm{\values}^2}{\numobs} +
\empnorm{\ftil - \fstar}^2 \Big \}
\end{align*}
for some universal constant $c'$. Taking expectations over
$\{\state_i\}_{i=1}^\numobs$ then yields
\begin{align}
\label{eqn:subgauss-control-expectation}
\EE \munorm{\fhat - \ftil}^2 \leq c' \Big \{ \EE
\left[\subgausscrit^2\right] + \frac{\inftynorm{\values}^2}{\numobs} +
\munorm{\ftil - \fstar}^2 \Big \}.
\end{align}
Finally,
combining~\cref{eqn:emp-control-expectation,eqn:subgauss-control-expectation},
we conclude that
\begin{align*}
\EE \munorm{\fhat - \ftil}^2 \leq c^{''} \Big \{
\EE\left[\subgausscrit^2\right] + \empcrit^2 + \frac{1}{\numobs}
\left(\bound^2 + \inftynorm{\values}^2 \right) + \munorm{\ftil -
  \fstar}^2 \Big \}.
\end{align*}
The theorem then follows from
applying~\Cref{lems:gausscrit-expectation} and rearranging.

\subsection{Proof of auxiliary lemmas for~\Cref{thm:regression}}
\label{sec:reg-lem-proofs}

We collect here the proofs of the auxiliary lemmas.

\subsubsection{Proof of~\Cref{lems:subgauss-local}}
\label{sec:proof-subgauss}

We begin with some straightforward algebraic manipulations. By
expanding the square, we have
\begin{multline*}
\frac{1}{\numobs} \sum_{i=1}^\numobs \left(\response_i - f(\state_i)
\right)^2 = \frac{1}{\numobs} \sum_{i=1}^{\numobs} \left( \ytil_i -
f(\state_i) \right)^2 + \frac{2}{\numobs} \sum_{i=1}^\numobs
\left(\fstar(\state_i) - \ftil(\state_i) \right) (\ytil_i -
f(\state_i)) + \frac{1}{\numobs} \sum_{i=1}^\numobs
\left(\fstar(\state_i) - \ftil(\state_i)\right)^2,
\end{multline*}
where we define $\ytil_i \defn \ftil(\state_i) + \noise_i$. Since the
rightmost term does not involve $f$, we can write
\begin{align*}
\fhat \in \arg \min_{f \in \fclass} \left\{ \frac{1}{\numobs}
\sum_{i=1}^\numobs \left(\ytil_i - f(\state_i) \right)^2 +
\frac{2}{\numobs} \sum_{i=1}^\numobs \left( \fstar(\state_i) -
\ftil(\state_i) \right) (\ytil_i - f(\state_i)) \right\}.
\end{align*}
By optimality of $\fhat$ and feasibility of $\ftil$, we have
\begin{multline*}
 \frac{1}{\numobs} \sum_{i=1}^\numobs \left(\ytil_i - \fhat(\state_i)
 \right)^2 + \frac{2}{\numobs} \sum_{i=1}^\numobs \left(
 \fstar(\state_i) - \ftil(\state_i) \right) (\ytil_i -
 \fhat(\state_i)) \\\leq \frac{1}{\numobs} \sum_{i=1}^\numobs
 \left(\ytil_i - \ftil(\state_i) \right)^2 + \frac{2}{\numobs}
 \sum_{i=1}^\numobs \left( \fstar(\state_i) - \ftil(\state_i) \right)
 (\ytil_i - \ftil(\state_i)).
\end{multline*}
Rearranging terms yields the \textit{basic inequality}
\begin{align}
\begin{split}
\label{eqn:basic-inequality}
\frac{1}{2\numobs} \sum_{i=1}^\numobs \left( \fhat(\state_i) -
\ftil(\state_i) \right)^2 &\leq \Term_1 + \Term_2,
\end{split}
\end{align}
where we define
\begin{align*}
\Term_1 \defn \frac{1}{\numobs} \sum_{i=1}^\numobs \noise_i \left(
\fhat(\state_i) - \ftil(\state_i) \right), \quad \text{and} \quad
\Term_2 \defn \frac{1}{\numobs} \sum_{i=1}^\numobs \left(
\fstar(\state_i) - \ftil(\state_i) \right) \left(\fhat(\state_i) -
\ftil(\state_i) \right).
\end{align*}
We bound each of $\Term_1$ and $\Term_2$ in turn.

\paragraph{Bounding $\Term_2$:} By applying the Fenchel-Young
inequality, we find
\begin{align*}
\Term_2 & \leq \frac{1}{\numobs} \sum_{i=1}^\numobs \left(
\fstar(\state_i) - \ftil(\state_i) \right)^2 + \frac{1}{4\numobs}
\sum_{i=1}^\numobs \left( \fhat(\state_i) - \ftil(\state_i) \right)^2
\\
& = \empnorm{\fstar - \ftil}^2 + \frac{1}{4} \empnorm{\fhat -
  \ftil}^2.
\end{align*}

\paragraph{Bounding $\Term_1$:} For this part, we condition on $\{ \state_i \}_{i=1}^\numobs$, 
i.e., we treat them as fixed and given to us. We make use of techniques for controlling
localized empirical processes; see Chapter 13 in the
book~\cite{dinosaur2019} for relevant background. Recall that by
definition, the scalar $\subgausscrit > 0$ satisfies the critical
inequality $\frac{\subgausscomp(\delta; \shiftfclass)}{\delta} \leq
\frac{\delta}{2}$. Recall that $\shiftfclass = \{ f - \ftil : f \in
\fclass \}$ is the shifted function class around $\ftil$. For a scalar
$u \geq \subgausscrit$, define the event
\begin{align}
\event(u) = \left\{ \text{ $\exists \,g \in \shiftfclass$ with
  $\empnorm{g} \geq u$ such that $\left| \frac{1}{\numobs}
  \sum_{i=1}^\numobs \noise_i \, g(\state_i) \right| \geq 2
  \empnorm{g} u$}\right\}.
\end{align}
The following result controls the probability of this event:
\begin{lemma}
\label{lems:subgauss-control}
For all $u \geq \subgausscrit$, we have
\begin{align*}
\PP(\event(u)) \leq e^{-c\numobs \cdot
  \frac{u^2}{\inftynorm{\values}^2}}
\end{align*}
for some universal constant $c > 0$.
\end{lemma}
\noindent
See~\Cref{sec:proof-subgauss-control} for the proof of this lemma. \\

\medskip

By~\Cref{lems:subgauss-control}, since $\fhat - \ftil \in
\shiftfclass$, we have if $\empnorm{\fhat - \ftil} > \sqrt{t
  \subgausscrit}$ for some $t \geq \subgausscrit$, then
\begin{align*}
|\Term_1| = \left| \frac{1}{\numobs}\sum_{i=1}^\numobs \noise_i
(\fhat(\state_i) - \ftil(\state_i)) \right| \leq 2 \empnorm{\fhat -
  \ftil} \sqrt{t \subgausscrit}
\end{align*}
with probability exceeding $1 - e^{-cn \cdot
  \frac{t\subgausscrit}{\inftynorm{\values}^2}}$.

Putting together the pieces, we conclude that
\begin{align*}
\frac{1}{2} \empnorm{\fhat - \ftil}^2 \leq \frac{1}{2} t\subgausscrit
+ 2 \empnorm{\fhat - \ftil}\sqrt{t\subgausscrit} + \empnorm{\fstar -
  \ftil}^2 + \frac{1}{4} \empnorm{\fhat - \ftil}^2,
\end{align*}
which implies that
\begin{align*}
\empnorm{\fhat - \ftil}^2 \leq 2 t \subgausscrit + 8 \sqrt{t
  \subgausscrit} \empnorm{\fhat - \ftil} + 4 \empnorm{\fstar -
  \ftil}^2.
\end{align*}
This inequality implies that there are universal constants $c_1, c_2,
c_3$ such that
\begin{align}
\empnorm{\fhat - \ftil}^2 \leq c_1 t \subgausscrit + c_2
\empnorm{\ftil - \fstar}^2
\end{align}
with probability exceeding $1 - e^{-c_3 n \cdot
  \frac{t\subgausscrit}{\inftynorm{\values}^2}}$.

\subsubsection{Proof of~\Cref{lems:gausscrit-expectation}}
\label{sec:proof-gausscrit-expectation}

%
It suffices to show that for any $u \geq \max\{\popgausscrit,
\frac{\empcrit^2}{\popgausscrit}\}$, we have the bound
\begin{align}
\label{eqn:SauteedScallops}  
  \subgausscrit^2 \leq 4 u \popgausscrit \qquad \mbox{with probability at
    least $1 - c_1 e^{-c_2 \numobs
      \frac{u\popgausscrit}{(\inftynorm{\values} + \bound)^2}}$.}
\end{align}
Indeed, the statement of the lemma can be obtained by integrating this
tail bound.  Accordingly, the remainder of our proof is devoted to
establishing inequality~\eqref{eqn:SauteedScallops} for some fixed but
arbitrary $u$.

Recall that by the definition~\eqref{eqn:critical}, the critical radii
satisfy the inequalities
\begin{align*}
\frac{\subgausscomp_\numobs(\subgausscrit,
  \shiftfclass)}{\subgausscrit} \leq \frac{\subgausscrit}{2}, \qquad
\frac{\subgausspop_\numobs(\popgausscrit,
  \shiftfclass)}{\popgausscrit} \leq \frac{\popgausscrit}{2}, \quad
\text{and} \quad \frac{\radcomp_\numobs(\empcrit;
  \shiftfclass)}{\empcrit} \leq \frac{\empcrit}{\bound}.
\end{align*}
For each $t$, define the random variable
\begin{align}
\label{eqn:empproccov-def}
\empproccov(t) \defn \EE_{\noise}\left[ \sup_{g \in \shiftfclass,
    \munorm{g} \leq t} \left| \frac{1}{\numobs} \sum_{i=1}^\numobs
  \noise_i g(\state_i) \right| \right],
\end{align}
and observe that $\EE[\empproccov(t)] = \radcomp_\numobs(t;
\shiftfclass)$ by construction. We also define the events
\begin{align}
\label{eqn:event-defn}
\event_0(t) \defn \left\{ \empproccov(t) \leq \radcomp_\numobs(t;
\shiftfclass) + \frac{t \sqrt{u \popgausscrit}}{8} \right\}, \quad
\text{and} \quad \event_1(t) \defn \left\{ \sup_{f \in \shiftfclass}
\frac{\left| \empnorm{f}^2 - \munorm{f}^2 \right|}{\munorm{f}^2 + t^2}
\leq \frac{1}{2} \right\}.
\end{align}
Conditional on $\event_1(t)$, we have for all $f \in \shiftfclass$
\begin{align*}
\empnorm{f} &\leq \sqrt{\tfrac{3}{2} \munorm{f}^2 + \tfrac{1}{2} t^2}
\leq 2 \munorm{f} + t, \qquad \text{and} \\ \munorm{f} &\leq \sqrt{2
  \empnorm{f}^2 + t^2} \leq 2 \empnorm{f} + t.
\end{align*}
Therefore conditioned on $\event_1(\sqrt{u \popgausscrit})$,
\begin{align}
\label{eqn:subgausscrit-change}
\subgausscomp_\numobs(s) = \EE_\noise \left[ \sup_{g \in \shiftfclass,
    \empnorm{g} \leq s} \left| \frac{1}{\numobs} \sum_{i=1}^\numobs
  \noise_i g(\state_i) \right|\right] \leq \EE_\noise \left[ \sup_{g
    \in \shiftfclass, \munorm{g} \leq 2s + \sqrt{u\popgausscrit}}
  \left| \frac{1}{\numobs} \sum_{i=1}^\numobs \noise_i g(\state_i)
  \right| \right] = \empproccov(2s + \sqrt{u\popgausscrit}).
\end{align}

Conditioning on both $\event_0\left(9\sqrt{u \popgausscrit} \right)$
and $\event_1(\sqrt{u\popgausscrit})$, we have
\begin{align*}
\subgausscomp(4\sqrt{u \popgausscrit}) \stackrel{(i)}{\leq}
\empproccov(9\sqrt{u\popgausscrit}) \stackrel{(ii)}{\leq}
\radcomp_\numobs(9\sqrt{u\popgausscrit}) + \frac{9}{8} u
\popgausscrit,
\end{align*}
where step (i) follows from the bound~\eqref{eqn:subgausscrit-change},
and step (ii) follows from the event $\event_0(9\sqrt{u
  \popgausscrit})$. Since the function $\delta \mapsto
\frac{\radcomp_\numobs(\delta)}{\delta}$ is non-increasing, and
$9\sqrt{u\popgausscrit} \geq \popgausscrit$,
\begin{align*}
\frac{\radcomp_\numobs(9\sqrt{u\popgausscrit})}{9\sqrt{u\popgausscrit}}
\leq \frac{\radcomp_\numobs(\popgausscrit)}{\popgausscrit} \leq
\frac{\popgausscrit}{2} \leq \frac{\sqrt{u\popgausscrit}}{2} \implies
\radcomp_\numobs(9\sqrt{u\popgausscrit}) \leq \frac{9}{2} u
\popgausscrit,
\end{align*}
using the definition of $\popgausscrit$ and the critical
inequality. We then have
\begin{align*}
\subgausscomp(4\sqrt{u\popgausscrit}) \leq \frac{9}{2} u \popgausscrit
+ \frac{9}{8} u \popgausscrit = \ \frac{45}{8} u \popgausscrit,
\end{align*}
which implies
\begin{align*}
\frac{\subgausscomp(4\sqrt{u\popgausscrit})}{4\sqrt{u\popgausscrit}}
\leq \frac{45}{32} \sqrt{u \popgausscrit} \leq \frac{4
  \sqrt{u\popgausscrit}}{2}.
\end{align*}
Thus $4\sqrt{u\popgausscrit}$ satisfies the critical inequality
defined by $\subgausscomp$, implying that $\subgausscrit \leq 4
\sqrt{u \popgausscrit}$, as desired.

The following lemma provides control on the events $\event_0(9\sqrt{u
  \popgausscrit})^c$ and $\event_1(\sqrt{u \popgausscrit})^c$.
\begin{lemma}
\label{lems:subgauss-change-prob}
For any $u \geq \max\{\popgausscrit, \tfrac{\empcrit^2}{\popgausscrit}
\}$, we have
\begin{align}
\PP(\event_0(9\sqrt{u \popgausscrit})^c) &\leq 2 \exp\left(-c_1
\numobs \cdot \frac{u \popgausscrit}{\inftynorm{\values}^2 + \bound
  \inftynorm{\values}} \right), \quad
\text{and} \label{eqn:event-0-control} \\ & \PP(\event_1(\sqrt{u
  \popgausscrit})^c) \leq c_2 \exp\left( -c_3 \numobs \cdot
\frac{u\popgausscrit}{b^2} \right). \label{eqn:event-1-control}
\end{align}
\end{lemma}
\noindent
See~\Cref{sec:proof-subgauss-change-prob} for the proof of this lemma.


\section{Discussion}
\label{sec:discussion}

In this paper, we presented and analyzed the Krylov--Bellman boosting
(KBB) procedure.  It is an efficient algorithm for sample-based policy
evaluation, and applicable to general state spaces.  It is built upon
the machinery of Krylov subspace methods for solving linear operator
equations.  As opposed to an idealized Krylov method, our approach
allows for errors in both the LSTD steps and the fitting of the
Bellman residual, and we provide general convergence guarantees that
track the effect of such errors.  On the empirical front, the KBB
algorithm typically exhibits convergence rates that are significantly
faster than the linear or geometric rate obtained by fitted value
iteration.  Consistent with these empirical observations, our theory
also reveals such accelerated convergence.  In particular, we show how
the KBB convergence rate is determined by the spectral properties of
the Bellman residual operator when restricted to a shrinking sequence
of subspaces.  Finally, we analyzed the use of non-parametric
regression routines for fitting the Bellman residual.  We proved a
general theorem that bounds the residual fitting error in terms of a
combination of approximation error, and statistical estimation error.

There exist many interesting avenues for future work. While the
current theory applies to reversible Markov chains, the method itself
exhibits rapid convergence for general Markov reward processes.  It
would be interesting to close this gap between theory and practice by
developing theory applicable to non-reversible Markov chains.  It is
also interesting to draw connections to related problems in
reinforcement learning and under different sampling models For
instance, can we fit the Bellman residuals using offline data, and
still provide reasonable guarantees?  How do updates of the
Krylov--Bellman type interact with actor-critic style algorithms?

\subsection*{Acknowledgements}  
This work was partially supported by Office of Naval Research Grant
ONR-N00014-21-1-2842, NSF-CCF grant 1955450, and NSF-DMS grant 2015454
to MJW. In addition, EX is supported by an NSF GRFP.

\bibliography{bibliography} \bibliographystyle{amsalpha}

\appendix


\section{Further Proofs}

Here we present the proofs of various lemmas used in the main text of
the paper.


\subsection{Proof of~\Cref{eqn:operator}}
\label{sec:proof-lemma-operator}

Let us state the result to be proven more formally:
\begin{lemma}
\label{lems:operator}
For any separable Hilbert space $\LsqClass$, the operator $(\IdOp -
\contractPar \TransOp)$ is self-adjoint, invertible, and satisfies the
bounds
\begin{align}
  (1 - \contractPar) \munorm{f}^2 \leq \statinprod{f}{(\IdOp -
    \contractPar \TransOp) f} \leq (1 + \contractPar) \munorm{f}^2
  \qquad \text{for all $f \in \LsqClass$.}
\end{align}
\end{lemma}

The requirements for $\LsqClass$ to be separable is quite mild; in the
case where $\stateset = \RR^d$ for some dimension $d$, one sufficient
condition is that $\stationary$ is a continuous distribution.

\begin{proof}

By definition of the stationary measure, the operator $\TransOp$
satisfies the bound $\munorm{\TransOp \values} \leq \munorm{\values}$.
Thus, by classical functional analysis (e.g., see Chap. 8 in the
reference~\cite{kreyszig1991introductory}), the inverse $(\IdOp -
\contractPar \TransOp)^{-1}$ exists, and can be represented via the
Neumann series expansion
\begin{align*}
(\IdOp - \contractPar \TransOp)^{-1} = \sum_{j=0}^\infty
  \contractPar^j \TransOp^j.
\end{align*}
The other statements follow from using linearity of the inner product,
and then applying Cauchy-Schwarz in conjunction with the operator norm
bound of $\TransOp$.

Since $\opQ \defn \IdOp - \contractPar \TransOp$ is a positive
operator, it has a unique square root $(\IdOp - \contractPar
\TransOp)^{1/2}$. In addition, since $\opQ$ is invertible, so is its
square root (which is also self-adjoint).
\end{proof}


\subsection{Proof of~\Cref{lems:subgauss-control}}
\label{sec:proof-subgauss-control}

This argument is inspired by the proof of Lemma 13.12 in the
book~\cite{dinosaur2019}, along with some new ingredients. We treat
the observations $\{ \state_i \}_{i=1}^\numobs$ as fixed here. Recall the definition of the event
\begin{align*}
\event(u) = \left\{ \text{ $\exists \,g \in \shiftfclass$ with
  $\empnorm{g} \geq u$ such that $\left| \frac{1}{\numobs}
  \sum_{i=1}^\numobs \noise_i \, g(\state_i) \right| \geq 2
  \empnorm{g} u$}\right\}.
\end{align*}
 Define
the empirical process
\begin{align*}
\empproc(u) \defn \sup_{\substack{g \in \shiftfclass, \\ \empnorm{g}
    \leq u}} \left| \frac{1}{\numobs} \sum_{i=1}^\numobs \noise_i \,
g(\state_i) \right|.
\end{align*}
We claim that
\begin{align*}
\PP\left( \event(u) \right) \leq \PP(\empproc(u) \geq 2 u^2).
\end{align*}
In order to establish this claim, suppose that there exists $g \in
\shiftfclass$ with $\empnorm{g} \geq u$ such that
\begin{align*}
\left| \frac{1}{n} \sum_{i=1}^\numobs \noise_i \, g(\state_i) \right|
\geq 2 \empnorm{g} u.
\end{align*}
Define the rescaled function $\gtil = \frac{u}{\empnorm{g}} g$;
observe $\gtil \in \shiftfclass$ by convexity.  We then have
\begin{align*}
\left| \frac{1}{\numobs} \sum_{i=1}^\numobs \noise_i \,
\gtil(\state_i) \right| = \frac{u}{\empnorm{g}} \left| \frac{1}{n}
\sum_{i=1}^\numobs \noise_i \, g(\state_i) \right| \geq 2u^2,
\end{align*}
as desired.

It remains to control the probability $\PP(\empproc(u) \geq 2u^2)$. If
we view $\empproc(u)$ as a function of $(\noise_1, \ldots,
\noise_\numobs)$, it can then be seen that it has Lipschitz constant
of $\frac{u}{\sqrt{n}}$ in the Euclidean norm and is separately
convex. Since $|\noise_i| \leq 2 \inftynorm{\values}$, we have by
Theorem 3.4 in the book~\cite{dinosaur2019},
\begin{align*}
\PP(\empproc(u) \geq \EE[\empproc(u)] + s) \leq e^{-cn \cdot \frac{s^2}{u^2 \inftynorm{\values}^2}}.
\end{align*}
Observe by definition that $\EE[\empproc(u)] = \subgausscomp(u)$, so
\begin{align*}
\frac{\subgausscomp(u)}{u} \stackrel{(i)}{\leq}
\frac{\subgausscomp(\subgausscrit)}{\subgausscrit}
\stackrel{(ii)}{\leq} \frac{\subgausscrit}{2} \leq \subgausscrit,
\end{align*}
where step~(i) uses Lemma 13.6 in the book~\cite{dinosaur2019} and the
fact that $u \geq \subgausscrit$, and step~(ii) uses the critical
inequality~\eqref{eqn:critical}. Thus, we conclude that
\begin{align*}
\PP(\empproc(u) \geq 2u^2) \leq \PP(\empproc(u) \geq u \subgausscrit +
u^2) \leq e^{-c\numobs \cdot \frac{u^2}{\inftynorm{\values}^2}},
\end{align*}
as desired.

\subsection{Proof of Lemma~\ref{lems:subgauss-change-prob}}
\label{sec:proof-subgauss-change-prob}

To begin, we observe that the claim~\eqref{eqn:event-1-control}
follows by applying~\Cref{lems:rad-local}, along with the fact that $u
\geq \tfrac{\empcrit^2}{\popgausscrit}$ implies that $\sqrt{u
  \popgausscrit} \geq \empcrit$.  In order to prove the
claim~\eqref{eqn:event-0-control}, assume that $u \geq \popgausscrit$,
and define the random variable
\begin{align*}
\empproc(t) \defn \sup_{g \in \shiftfclass, \munorm{g} \leq t} \left|
\frac{1}{\numobs} \sum_{i=1}^{\numobs} \noise_i \, g(\state_i)
\right|.
\end{align*}
By the definition~\eqref{eqn:empproccov-def} of $\empproccov(t)$, we
have the equivalence
\begin{align*}
\empproccov(t) = \EE_\noise \left[ \empproc(t) \right] =
\EE[\empproc(t) \mid \{\state_i\}_{i=1}^\numobs],
\end{align*}
and $\EE\left[\empproccov(t)\right] = \EE[\empproc(t)] =
\subgausspop_\numobs(t)$. The moment generating function of
$\empproccov(t) - \subgausspop_\numobs(t)$ can be bounded as
\begin{align*}
\EE\left[e^{\lambda(\empproccov(t) - \subgausspop_\numobs(t))} \right]
= \EE\left[e^{\lambda(\EE_\noise[\empproc(t)] -
    \subgausspop_\numobs(t))} \right] \leq \EE\left\{ \EE_\noise
\left[e^{\lambda(\empproc(t) - \subgausspop_\numobs(t))} \right]
\right\} = \EE\left[ e^{\lambda(\empproc(t) -
    \subgausspop_\numobs(t))} \right],
\end{align*}
where the inequality follows from a (conditional) Jensen's
inequality. Consequently, we can control $\empproccov(t) -
\subgausspop_\numobs(t)$ by controlling $\empproc(t) -
\subgausspop_\numobs(t)$ using a Chernoff-type argument. We prove an
upper tail bound on $\empproc(t)$ via Talagrand
concentration~\cite{talagrand1996funcbern} which uses a Chernoff bound
in its proof; see Theorem 3.27 in the book~\cite{dinosaur2019} for
further details.

Define the random variable $\empsquare = \sup_{f \in \shiftfclass, \munorm{f} \leq t} \frac{1}{\numobs} \sum_{i=1}^\numobs \noise_i^2 f(\state_i)^2$. We then have by Talagrand concentration
\begin{align*}
\PP(\empproc(t) \geq \subgausspop_\numobs(t) + s) \leq 2 \exp \left(-c \numobs \cdot \frac{s^2}{\EE[\empsquare] + \bound s} \right)
\end{align*}
where $c$ is a universal constant. By Equation (3.84) from the book~\cite{dinosaur2019}, we have
\begin{align*}
\EE[\empsquare] &\leq \sup_{f \in \shiftfclass, \munorm{f} \leq t} \EE[\noise^2 f^2(\state)] + 4 b \inftynorm{\values} \cdot \subgausspop_\numobs(t) \\
& \leq 4 \inftynorm{\values}^2 \, t^2 + 4 b \inftynorm{\values} \cdot \subgausspop_\numobs(t),
\end{align*}
where we have used the fact $|w| \leq 2 \inftynorm{\values}$ a.s. Thus, we have
\begin{align*}
\PP\left(\empproccov(9\sqrt{u\popgausscrit}) \geq \subgausspop_\numobs(9\sqrt{u\popgausscrit}) + \frac{9u\popgausscrit}{8}   \right) &\leq 2\exp\left(-c \numobs \cdot \frac{u^2 \popgausscrit^2}{\inftynorm{\values}^2 u \popgausscrit + b \inftynorm{\values} \cdot \subgausspop_\numobs(9\sqrt{u\popgausscrit}) + b\inftynorm{\values} u\popgausscrit} \right) \\
&\stackrel{(i)}{\leq} 2 \exp\left(-c' \numobs \cdot \frac{u \popgausscrit}{\inftynorm{\values}^2 + \bound \inftynorm{\values}}\right),
\end{align*}
where step (i) follows from the fact that $\subgausspop(9\sqrt{u\popgausscrit}) \leq \frac{9}{2} u\popgausscrit$ for $u \geq \popgausscrit$.


\section{Calculations for~\Cref{sec:reg-further-details}}
\label{sec:critical-computations}

In this section we provide the details behind the calculations in~\Cref{sec:reg-further-details}. Before proceeding to the computations, we have the following result relating the sub-Gaussian complexity $\subgausspop(\delta; \shiftfclass)$ and the Rademacher complexity $\radcomp(\delta; \shiftfclass)$
\begin{lemma}
\label{lems:subgauss-rad-comparison}
We have for some universal constant $c$,
\begin{align*}
\subgausspop(\delta; \shiftfclass) \leq c \inftynorm{\values} \cdot \radcomp(\delta; \shiftfclass).
\end{align*}
\end{lemma}
\noindent
This lemma allows us to instantly obtain a bound for the sub-Gaussian complexity after computing the Rademacher complexity of the function class $\shiftfclass$. In~\Cref{sec:reg-further-details}, we assume $\inftynorm{\values} = 1$; consequently $\empcrit^2 \lesssim \popgausscrit^2$.

\begin{proof}
We have by the tower property,
\begin{align*}
\subgausspop(\delta; \shiftfclass) &= \EE\left[ \EE_\noise \left[ \sup_{g \in \shiftfclass, \statnorm{g} \leq \delta} \left| \frac{1}{\numobs} \sum_{i=1}^\numobs \noise_i g(\state_i) \right|\right] \right].
\end{align*}
We can then bound, letting $\rad_i$ be i.i.d.~Rademacher random variables independent of $\covariateset$,
\begin{align*}
\EE_\noise \big[ \sup_{g \in \shiftfclass, \statnorm{g} \leq \delta} \big| \frac{1}{\numobs} \sum_{i=1}^\numobs \noise_i g(\state_i) \big|\big] &\stackrel{(a)}{\leq} 2\EE_{\noise, \rad}\big[ \sup_{g \in \shiftfclass, \statnorm{g} \leq \delta} \big| \frac{1}{\numobs} \sum_{i=1}^\numobs \rad_i \noise_i g(\state_i) \big| \big] \\
&= 4\inftynorm{\values} \cdot \EE_{\noise, \rad}\left[ \sup_{g \in \shiftfclass, \statnorm{g} \leq \delta} \left| \frac{1}{\numobs} \sum_{i=1}^\numobs \rad_i \cdot \frac{\noise_i g(\state_i)}{2\inftynorm{\values}} \right| \right] \\
&\stackrel{(b)}{\leq} 8 \inftynorm{\values} \cdot \EE_{\rad} \big[ \sup_{g \in \shiftfclass, \statnorm{g} \leq \delta} \big| \frac{1}{\numobs} \sum_{i=1}^\numobs \rad_i g(\state_i) \big|\big],
\end{align*}
where step~(a) follows from symmetrization (cf. Chapter 4 in the reference~\cite{dinosaur2019}) and step~(b) follows from the bound $|\noise_i| \leq 2\inftynorm{\values}$ and the Ledoux-Talagrand contraction inequality~\cite{ledoux1991probability}.

\end{proof}

\subsection{Polynomials of degree $\polydeg$}
\label{sec:poly-class-critical}

To simplify the calculations, let $\basis_0, \ldots, \basis_d$ be an
orthonormal basis of $\shiftfclass$ with respect to
$\statinprod{\cdot}{\cdot}$. This basis exists via applying
Gram-Schmidt to the basis $\{1, x, \ldots, x^d\}$. We can then write
any polynomial function in $\shiftfclass$ as $f_\orthcoeffs(x) =
\orthcoeffs_0 \basis_0(x) + \orthcoeffs_1 \basis_1(x) + \ldots +
\orthcoeffs_d \basis_d(x)$, which satisfies $\statnorm{f_\orthcoeffs}
= \twonorm{\orthcoeffs}$. Define the matrix $\evalmat \in \RR^{\numobs
  \times (\polydeg + 1)}$ with $\evalmat_{ij} =
\basis_j(\state_i)$. We then have, letting (with mild abuse of
notation) $\rad = (\rad_1, \ldots, \rad_\numobs) \in \RR^{\numobs}$,
\begin{align*}
\subgausspop_\numobs(\delta; \shiftfclass) &= \EE\left[\sup_{f_\orthcoeffs \in \shiftfclass, \statnorm{f_\orthcoeffs} \leq \delta} \left| \frac{1}{\numobs} \sum_{i=1}^\numobs \rad_i f_\orthcoeffs(\state_i) \right| \right] \\
&\leq \EE\left[ \sup_{\twonorm{\orthcoeffs} \leq \delta} \left|\frac{1}{\numobs} \rad^T \evalmat \orthcoeffs \right|\right] \stackrel{(a)}{=} \frac{\delta}{\numobs} \EE\left[ \twonorm{\rad^T \evalmat} \right] \stackrel{(b)}{\leq} \sqrt{\EE[\twonorm{\rad^T \evalmat}^2]},
\end{align*}
where step~(a) follows from the fact that $\twonorm{\cdot}$ is a dual norm to itself, and step~(b) follows from Jensen's inequality. 

Manipulating further, we have
\begin{align*}
\EE\left[ \twonorm{\rad^T \evalmat}^2 \right] = \EE\left[ \EE_\rad[\text{trace}(\evalmat^T \rad \rad^T \evalmat)] \right] = \EE\left[\text{trace}(  \EE_\rad[\rad \rad^T] \evalmat \evalmat^T) \right],
\end{align*}
where we used the fact $\EE_\rad[\cdot] \defn \EE[\cdot \mid \{\state_i \}_{i=1}^\numobs]$ and $\evalmat$ is a function of $\covariateset$. We then have
\begin{align*}
\EE\twonorm{\rad^T \evalmat}^2 \stackrel{(a)}{=} \EE[\text{trace}(\evalmat \evalmat^T)] \stackrel{(b)}{=} (\polydeg + 1)\numobs,
\end{align*}
where the equality~(a) follows from the fact that $\EE[\rad\rad^T] = \Id$, and the equality~(b) follows from the fact that $\{\basis_0, \ldots, \basis_d \}$ is orthonormal. To identify the critical radius, we compute using the definition~\eqref{eqn:critical},
\begin{align*}
\frac{4\delta \sqrt{\polydeg+1}}{\delta \sqrt{\numobs}} \leq \delta \implies \empcrit^2 \lesssim \frac{d+1}{\numobs}. 
\end{align*}
By~\Cref{lems:subgauss-rad-comparison}, we have $\popgausscrit^2 \lesssim \frac{d+1}{\numobs}$.

\subsection{Radial basis kernels}

By Mercer's theorem (see Chapter 12 in the book~\cite{dinosaur2019}), we have the kernel integral operator $T_\kernel(f)(\state) \defn \int_{\stateset} \kernel(\state, z) f(z) \, d \stationary(z)$ has a set of eigenvalues $\{\mu_{j}\}_{j=1}^\infty$. Then we have (cf. Corollary 14.5~\cite{dinosaur2019})
\begin{align*}
\radcomp_\numobs(\delta; \shiftfclass) \leq \sqrt{\frac{2}{\numobs}} \cdot \sqrt{\sum_{j=1}^\infty \min\{ \mu_j, \delta^2 \}}.
\end{align*}
Since a reversible Markov chain has a uniform stationary distribution, the eigenvalues of $T_\kernel$ satisfies
\begin{align*}
\mu_j \leq c_0 e^{-c_1 j \log j} \qquad \text{for $j= 1, 2, \ldots$}
\end{align*}
for some universal constants $c_0, c_1$ (see Chapter 12 in the reference~\cite{dinosaur2019}).

Let $k$ be the smallest positive integer such that $c_0 e^{-c_1 k \log k} \leq \delta^2$. We have
\begin{align*}
\sqrt{\frac{2}{\numobs}} \cdot \sqrt{\sum_{j=1}^\infty \min\{ \mu_j, \delta^2 \}} &\leq \sqrt{\sum_{j=1}^\infty \min\{\delta^2, c_0e^{-c_1 j \log j}\}} \\
&\leq \sqrt{k\delta^2 + c_0 \sum_{j=k+1}^\infty e^{-c_1 j \log j}} \\
&\leq \sqrt{k \delta^2 + c_0 e^{-c_1 k \log k}} \leq c_0 \sqrt{k \delta^2},
\end{align*}
where $c_0, c_1$ denote universal constants whose values may change from line to line. Some algebra then shows $\empcrit^2 \lesssim \frac{\log(\numobs + 1)}{\numobs}$ and $\popgausscrit^2 \lesssim \frac{\log(\numobs + 1)}{\numobs}$, as claimed.

\subsection{Convex Lipschitz functions} 
We use results involving chaining and empirical process theory to compute the critical radii in this setting, see the references~\cite{dinosaur2019, vandeGeer, talagrand2014upper} for a full exposition. Recall the definition of the event $\event_1(t)$ from~\Cref{eqn:event-defn}, and let $\empcrit$ denote the critical radius. By an argument similar to that of the proof of~\Cref{lems:gausscrit-expectation}, we have for $k \geq 1$,
\begin{align*}
\EE_{\rad} \left[\sup_{g \in \shiftfclass, \statnorm{g} \leq \empcrit} \left| \frac{1}{\numobs} \sum_{i=1}^{\numobs} \rad_i \, g(\state_i) \right|\right] &\leq \EE_{\rad} \left[\sup_{g \in \shiftfclass, \empnorm{g} \leq (k+2)\empcrit} \left| \frac{1}{\numobs} \sum_{i=1}^{\numobs} \rad_i \, g(\state_i) \right| \right],
\end{align*}
with probability exceeding $1 - c_0 e^{-c_1 \numobs \frac{k^2 \empcrit^2}{b^2}}$. We bound the right-hand side via chaining: we know from existing results (see Chapter 14 in the book~\cite{dinosaur2019}) that for Lipschitz function class $\shiftfclass$,
\begin{align*}
\EE_{\rad} \left[\sup_{g \in \shiftfclass, \empnorm{g} \leq (k+2)\empcrit} \left| \frac{1}{\numobs} \sum_{i=1}^{\numobs} \rad_i \, g(\state_i) \right| \right] \leq \frac{1}{\sqrt{\numobs}} \left((k+2) \, \empcrit\right)^{3/4}.
\end{align*}

Defining
\begin{align*}
Z \defn \EE_{\rad} \left[\sup_{g \in \shiftfclass, \empnorm{g} \leq (k+2)\empcrit} \left| \frac{1}{\numobs} \sum_{i= 1}^{\numobs} \rad_i \, g(\state_i) \right| \right],
\end{align*}
we can integrate the tail bound to get
\begin{align*}
\EE[Z] &= \int_0^\infty \PP(Z \geq u) \, du \\
&\leq \frac{1}{\sqrt{\numobs}} (3 \, \empcrit)^{3/4} + \frac{1}{\sqrt{\numobs}} \int_{(3\empcrit)^{3/4}}^\infty \PP(Z \geq u) \, du \\
&\stackrel{(a)}{\leq} \frac{1}{\sqrt{\numobs}} (3 \, \empcrit)^{3/4} + \frac{3c_0 \empcrit^{3/4}}{4} \int_0^\infty \exp\left(- c_1 \numobs k^2 \empcrit^2\right) \cdot (k+2)^{-1/4} \, dk \\
&\leq (3 \, \empcrit)^{3/4} + \frac{3c_0 \empcrit^{3/4}}{4\sqrt{\numobs}} \int_0^\infty \exp\left(- c_1 \numobs k^2 \empcrit^2\right)\, dk \\
&\stackrel{(b)}{=} \frac{1}{\sqrt{\numobs}}(3 \, \empcrit)^{3/4} + \frac{3c_0 \empcrit^{3/4}}{8\sqrt{n}} \cdot \sqrt{\frac{\pi}{n \empcrit^2}},
\end{align*}
where step~(a) follows from the change of variables $u = ((k+2) \, \empcrit)^{3/4}$ and step~(b) follows from the integral of a Gaussian density.  By definition~\eqref{eqn:critical}, we have $\radcomp_n(\empcrit; \shiftfclass) = \empcrit^2$, so we have for universal constants $c_1$ and $c_2$,
\begin{align*}
\empcrit^2 \leq c_1 \frac{\empcrit^{3/4}}{\sqrt{\numobs}} + \frac{c_2}{\empcrit^{1/4} \numobs}.
\end{align*}
Thus we conclude $\empcrit^2 \lesssim \frac{1}{n^{4/5}}$, as claimed.

\end{document}